\documentclass[10pt,twocolumn,letterpaper]{article}

\usepackage{cvpr}
\usepackage{times}
\usepackage{epsfig}
\usepackage{graphicx}
\usepackage{amsmath}
\usepackage{amssymb}

\usepackage{lipsum}
\usepackage{epigraph}
\usepackage{multirow}
\usepackage{booktabs}
\usepackage{tabularx}
\usepackage{colortbl}
\usepackage{epstopdf}
\usepackage{balance}
\usepackage{lib}
\usepackage{url}
\usepackage[font=small,labelfont=bf]{caption}
\usepackage{enumitem}

\newcommand{\vertical}[1]{\rotatebox[origin=l]{90}{\parbox{1.0cm}{#1}}}
\newcommand{\verticala}[1]{\rotatebox[origin=l]{90}{\parbox{3.5cm}{#1}}}
\newcolumntype{C}{>{\centering\arraybackslash}m{25ex}}
\newcommand\blfootnote[1]{
\begingroup \renewcommand\thefootnote{}\footnote{#1}
\addtocounter{footnote}{-1} \endgroup}
\newcommand{\fastrcnn}[0]{Fast R-CNN\xspace}

\usepackage[pagebackref=true,breaklinks=true,letterpaper=true,colorlinks,bookmarks=false]{hyperref}
\setlist[enumerate]{leftmargin=*}

\cvprfinalcopy % *** Uncomment this line for the final submission

\begin{document}

%%%%%%%%% TITLE
\title{Exploring Person Context and Local Scene Context for Object Detection}

\author{
\begin{tabular}{CCC}
Saurabh Gupta$^*$ & Bharath Hariharan$^*$ & Jitendra Malik\\
UC Berkeley & Facebook AI Research & UC Berkeley\\
{\tt\small sgupta@eecs.berkeley.edu} & {\tt\small bharathh@fb.com} & {\tt\small malik@eecs.berkeley.edu} 
\end{tabular}
}

\maketitle
\blfootnote{* Authors contributed equally}

%%%%%%%%% ABSTRACT
\begin{abstract}
In this paper we explore two ways of using context for object detection. The first model focusses on people and the objects they commonly interact with, such as fashion and sports accessories. The second model considers more general object detection and uses the spatial relationships between objects and between objects and scenes. Our models are able to capture precise spatial relationships between the context and the object of interest, and make effective use of the appearance of the contextual region. On the newly released COCO dataset, our models provide relative improvements of upto 5\% over CNN-based state-of-the-art detectors, with the gains concentrated on hard cases such as small objects (10\% relative improvement). 
\end{abstract}

\definecolor{Gray}{gray}{0.85}
\newcolumntype{g}{>{\columncolor{Gray}}c}

\section{Introduction}
\begin{figure}
\centering
\resizebox{1.0\linewidth}{!}{
\includegraphics[height=0.12\textheight]{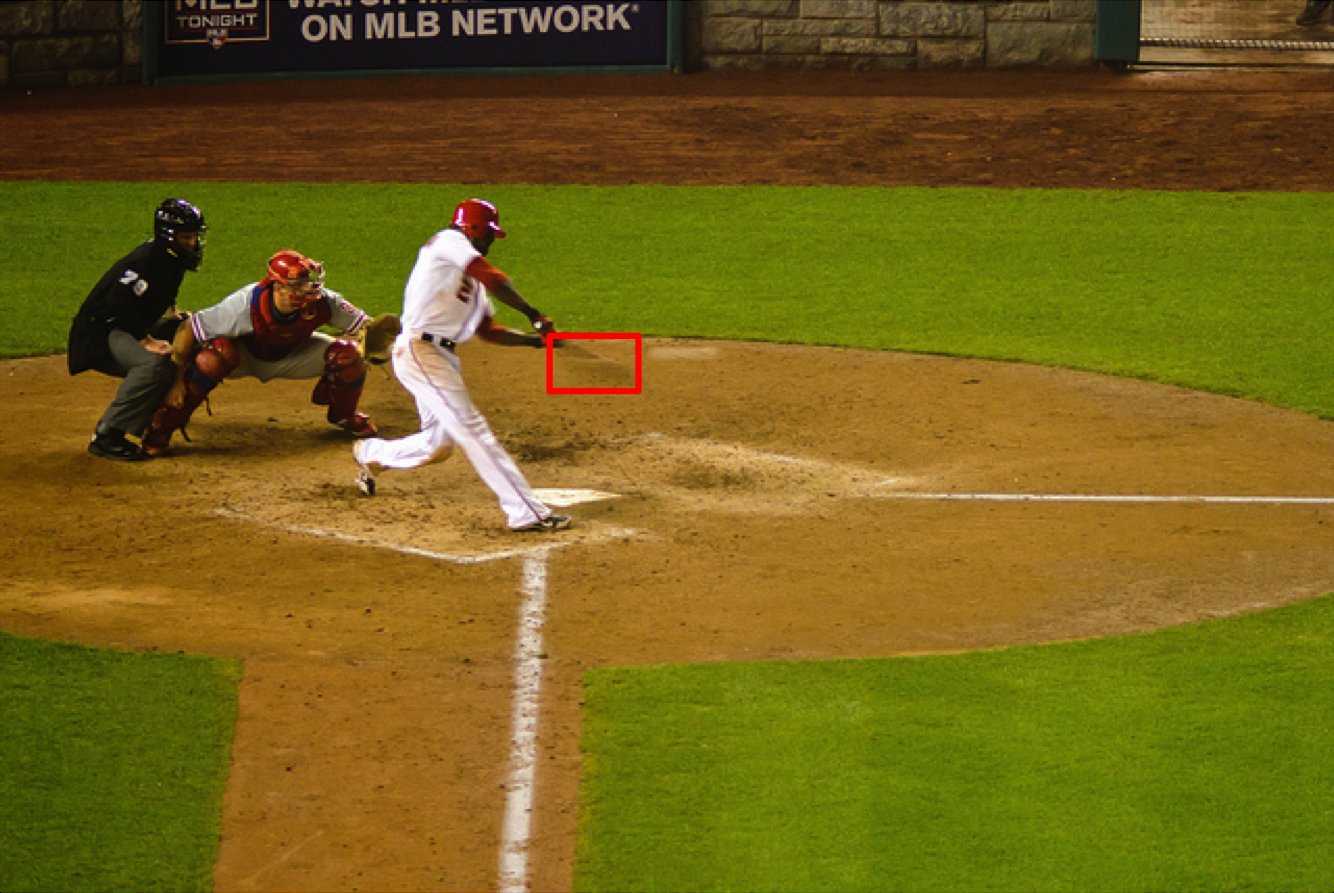}
\includegraphics[height=0.12\textheight]{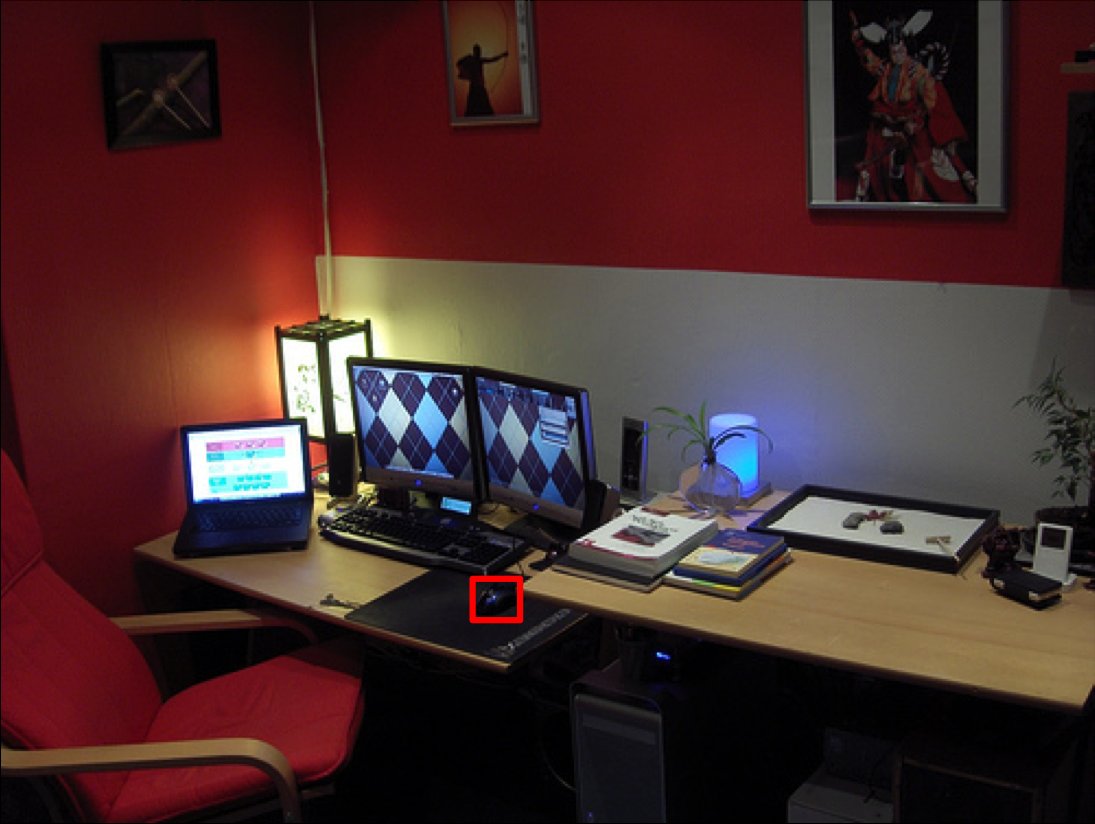}
}
\caption{Two examples of objects that are hard to detect except in context: a baseball bat and a mouse.}
\label{fig1}
\end{figure}

Current state-of-the-art object detectors are based on classifying region
proposals. Consider what such an approach would do with the baseball bat in the
left image in Figure~\ref{fig1}. The visual features on the bat are ambiguous
and indistinguishable from the background. Even the most sophisticated
classifier will be hard-pressed to classify it as a baseball bat. Instead,
consider an alternative strategy: we look at the people in the image, see if
any of them looks like a batter, locate his hands and search in the vicinity.
While the baseball bat itself seems to be indistinguishable from the
background, the person holding the bat is not only easily detectable but easily
identifiable as a batter, and his pose is distinctive and indicative of where
the bat should be.

Similar difficulties hinder the detection of the mouse in the right image.
Again, the mouse is barely distinguishable from its surroundings, and visual
features on the mouse itself do not lend themselves to easy classification. On
the contrary, the monitors are bright and easily discernible, as is the
keyboard. Once we find these two, it is easy to find the mouse to the side of
the keyboard and below the monitors.

These examples point towards an old idea in computer vision: ``context''.
However, compared to the typical instantiation of this idea, the reasoning described above is very sophisticated. First, the
``contextual'' region (the person in the first case or the monitor in the
second) is not fixed and can be far away from the object of interest.
Hard-wiring this context region to the whole global scene or to the immediate
neighborhood of the region being classified will likely not work. Second, the
spatial relationships that we want to infer are very precise: the batter
informs us not only of the presence of a baseball bat (as in a typical model
using image-level scene context), but also of its precise location and pose.
Third, this precise spatial relationship is conditioned on the appearance of
the contextual region. For instance, to detect the bat it is not enough to know
that there is a person in the vicinity. We need to look at the person, infer
that he is a batter, and determine his pose. 

In this paper we explore models that perform such reasoning. In particular, we
explore two models: 

\begin{enumerate}
\item Our first model focuses on people and on objects that have specific
relationships to people, such as the baseball bat above. We consider fashion
and sports accessories such as gloves, cellphones, handbags, ties, skis and
skateboards. We call this kind of context ``person context".  
\item Our second model considers general object detection and tries to use the
spatial relationship that objects typically hold with each other, such as the
relationship of a mouse with a keyboard and computer monitor. We call this kind
of context ``local scene context".
\end{enumerate}

We show that both person context and scene context offer up to 5\% relative
improvement for fashion and sports accessories. Local scene context also offers
similar gains in overall performance, modulo the upper bound set by object
proposal methods. These improvements, while small overall, are unevenly spread
across categories: our contextual models offer large gains ($>3$ points) for
several categories, especially those that current state-of-the-art object
detectors do badly on: baseball bats and gloves, mouse, kites etc. In addition
to improvements in detection, we are also able to link together people with
accessories that they are using. Such reasoning can be helpful for downstream
applications that want to go beyond object locations and aim to meaningfully
understand a scene. We will make the code public upon acceptance.
 
The rest of the paper is laid out as follows. In Section~\ref{sec:relwork} we
discuss prior work on context. After providing some background in
Section~\ref{sec:prelims}, we describe our two models in
Section~\ref{sec:personcontext} and Section~\ref{sec:scenecontext}. We end
with a conclusion.

\section{Related Work}
\label{sec:relwork}
Context has a long history in computer vision. Biederman~\cite{Biederman1981} proposed five different kinds of relationships between objects: support, interposition, probability, location and size. Our focus is on the latter three that deal with semantics: how likely certain objects are in particular locations and scales in a scene. Several researchers have put these ideas into practice. A lot of work has focused on using global scene features to identify possible locations of objects~\cite{RussellNIPS2007,oliva2007role, TorralbaIJCV2003, TorralbaICCV2001}. At the other extreme, capturing the immediate local neighborhood of the object by using a larger window for computing features has also proved useful, even with sophisticated classifiers~\cite{GidarisICCV2015,ZhuCVPR2015, MottaghiCVPR2014}. CRF-based approaches have tried to capture both the local consistency between adjacent pixels or detections and the consistency between detections and the entire scene~\cite{GonfausCVPR2010,YaoCVPR2012}. There is also work on reasoning about the 3D layout of objects, surface orientations etc.~\cite{Hoiem08}. 

More closely related to this paper is prior work on leveraging the relationship between objects. The simplest variants involve using the co-occurrence between object categories~\cite{Rabinovich07} or using the scores of detectors of other categories in the vicinity of a box as additional features~\cite{lsvm-pami}. However, this ignores spatial relationships between the objects. Desai \etal~\cite{desai2009discriminative} incorporate spatial relationships into this contextual scoring by framing it as a structured prediction problem. Such contextual rescoring is an instantiation of the more general idea of autocontext~\cite{TuCVPR2008}, an iterative procedure where predictions from a previous iteration are used to provide contextual features for the next iteration. Choi \etal~\cite{ChoiCVPR2010} reason both about the spatial relationship of object categories and about the global presence or absence of these categories, thus bringing in global scene context. These approaches suffer from the coarseness of category labels: whether there is a ski below a person depends on the pose of the person and on her appearance, variables that are marginalized over when converted to category-level scores. Yao and Fei-Fei~\cite{yao2010modeling} use the pose of people to help detect objects in a sports dataset. Our person context model is based on similar motivation, but builds on more powerful state-of-the-art object detection models~\cite{fastrcnn}.

A crucial issue in contextual models that operate at the level of category labels or pose is that they assume that only objects from the small pool of annotated categories can act as contextual information. This assumption isn't necessarily justified. There has been some recent work on automatically figuring out the right contextual region. Gkioxari \etal~\cite{GkioxariICCV2015} automatically pick contextual regions for action classification. Vezhnevets and Ferrari~\cite{VezhnevetsBMVC2015} use all boxes in an image to classify a bounding box. While these models are superficially similar to our local scene context model, the set of spatial relationships we capture is much richer. We are also able to use context from disparate regions of the image.Visual phrases~\cite{amin2011phrases, desai2012detecting} capture context by training detectors for pairs of objects that commonly co-occur. However, this requires not only knowing \apriori which pairs of categories to train detectors for but also training separate classifiers for each pair of categories, hurting generalization. Li \etal~\cite{LiICCV2011} remove the first restriction by training detectors for an object together with a large region surrounding it, but still have to deal with the loss in generalization caused by having multiple independent detectors.

Most prior work on context uses weak visual features such as HOG~\cite{Dalal05}. With weak features, context offers additional information that can help the classifier do a better job. However, object detection has now moved on to more powerful feature representations provided by convolutional networks~\cite{lecun-89e, krizhevsky2012imagenet, simonyan2014very, fastrcnn, szegedy2014going, szegedy2014scalable, zhang2015improving}.  It is an open question if reasoning about context still offers gains when working with CNNs. Gkioxari \etal~\cite{GkioxariICCV2015} do show that context helps, but the task is one of action classification rather than object detection. Zheng \etal~\cite{ZhengICCV2015}  showed that CRFs can be trained as recurrent networks, and use them to capture pixel-level context for semantic segmentation. Gidaris and Komodakis~\cite{GidarisICCV2015} and Zhu \etal~\cite{ZhuCVPR2015} show gains from looking at the immediate neighborhood of the box, but it is unclear if more sophisticated contextual reasoning is still necessary. One reason why we haven't seen context being more widely useful is that object detection benchmark datasets such as PASCAL VOC~\cite{PASCAL-ijcv} typically only contain large, distinctive object categories that stand out on their own. The recently released COCO dataset~\cite{mscoco} provides accurate and detailed annotations for a much larger set of object categories, including those which are small and hard to detect except in the light of their context. This dataset forms our testbed for our explorations on context.

\section{Preliminaries}
\label{sec:prelims}
Before delving into the details of our models, in this section we describe some useful background. The object detection pipelines we experiment with in this paper are based on the paradigm of producing region proposals and then classifying them with powerful classifiers. For region proposals we use MCG~\cite{arbelaezCVPR14} which is very effective at capturing objects from a large range of scales.

For classifying proposals, we build on top of Fast R-CNN~\cite{fastrcnn}. Fast R-CNN uses convolutional networks~\cite{lecun-89e} to classify region proposals. First, it extracts convolutional feature maps by passing the image (upsampled to a fixed size) through a convolutional network. Next, for each bounding box candidate, it uses an ROI Pooling layer to extract a fixed length feature vector from the convolutional feature maps. This layer superimposes a spatial grid with a fixed number of grid cells on top of the bounding box, and for each grid cell, for every channel, it records the maximum value of that feature channel in that grid cell (i.e, max pooling). This produces a fixed dimensional feature vector that is passed through two fully connected layers. The final feature vector is passed into a classifier that produces scores for each category, and a bounding box regressor that produces a new bounding box for each category. 

The entire CNN (including both the convolutional feature maps and the fully connected layers) is jointly trained using SGD. Each minibatch iteration samples 2 images and 64 boxes per image. Only boxes that overlap with a ground truth object by at least 10\% are considered for training. Those that overlap with a ground truth object by less than 50\% are labeled background, and those that overlap a ground truth object by more than 50\% are labeled with the corresponding class. A fourth of the boxes in each iteration are constrained to be positive. The bounding box regressor for a given category is only trained on the positive boxes for that category.  The loss for the classifier is negative log likelihood, while the loss for the bounding box regressor is the Huber loss:
\[
L_{\mathrm{H}} (\delta) = \left\{	\begin{array}{lr}
						\frac{\delta^2}{2} , & \mathrm{for } |\delta|<1 \\
						|\delta|-\frac{1}{2}, & \mathrm{ for } |\delta| \geq 1
						\end{array}\right.
\]
where $\delta$ is the residual error between the predicted bounding box and the ground truth. The convolutional network is initialized by pretraining on ImageNet classification.

In this paper we focus on two CNN architectures. The first architecture, which we call Alexnet, corresponds to the architecture used by Krizhevsky \etal~\cite{krizhevsky2012imagenet} and has five convolutional layers interspersed with max pooling layers followed by two fully connected layers. The second architecture, which we will refer to as VGG, is the 16-layer `D' network proposed by Simonyan and Zisserman~\cite{simonyan2014very}. For all our experiments we use the COCO dataset~\cite{mscoco}. We do all our training on the train set, and all our experiments on a 10000 image subset of the validation set (called val01 in the rest of the paper). Evaluation is done using the COCO code with a maximum of 100 detections per image. Test numbers are presented for the final models.

Fast R-CNN with VGG trained out-of-the-box on COCO train for 240000 iterations achieves a mean AP of $37.6$. For all our models, we start with this network as an initialization. To provide a fair comparison, we used this Fast R-CNN network to initialize another round of Fast R-CNN training, with 320000 more iterations, and with 500 boxes and 1 image per minibatch (to match the way we trained our proposed models)\footnote{Additional subtle changes are described in the supplementary material}. This leads to the network in general seeing more data, and gives a mean AP of 41.9. We use this number as the baseline in all our experiments.

\section{Person Context}
\label{sec:personcontext}
Our first model focuses on using person context to improve detection of objects that are usually linked to people. We call these objects ``person add-ons". We identify fifteen add-ons: backpack, umbrella, handbag, tie, suitcase, skis, snowboard, sports-ball, baseball-bat, baseball-glove, skateboard, surfboard, tennis-racket, wine-glass, and cell-phone. Our model consists of two steps:
\begin{enumerate}
\item We first get person detections from an off-the-shelf person detector (Fast R-CNN with Alexnet/VGG). For each of these person detections, for each add-on object category, we predict a score indicating whether there is an object of that category in the vicinity of the person, and a tentative location of this object.
\item We then assemble these person-centric predictions into a heatmap for every person add-on, where the score of a location indicates how likely it is that there is an object of that category at that location. We then consider these heatmaps as additional feature channels that are used by the object detector.
\end{enumerate}

We describe each step in detail below.
\subsection{Predicting person add-ons}
\label{subsec:personzoom}
We assume that each person is ``attached to'' at most one object of each add-on category. Given a person detection $d$, we use a convolutional network to predict, for every add-on category $i$, a probability $p_{id}$ indicating whether an object of that category is attached to the person, and a location $l_{id}$ for where the attached object might be. $l_{id}$ is parametrized as a 4-tuple $(\frac{\delta x}{\bar{w}}, \frac{\delta y}{\bar{h}}, \log\frac{w}{\bar{w}} , \log\frac{h}{\bar{h}})$ where $\delta x$ and $\delta y$ are displacements of the center of the add-on relative to the person center, $w,h$ are the dimensions of the add-on box, and $\bar{w},\bar{h}$ are the dimensions of the person.

Such reasoning requires analyzing the person in detail. This motivates two choices that we make:
\begin{enumerate}
\item In order to provide enough resolution, we do this reasoning on a high resolution view of the person. Concretely, we crop the bounding box of the person (expanded by a factor of $1.2$), reshape it to a fixed size ($227 \times 227$ for Alexnet, $224 \times 224$ for VGG) and feed it into the convolutional network.
\item We use an iterative scheme to predict the location. The first step uses the full person to predict a location $\hat{l}_{id}$ for every add-on category $i$. The target for this step is a box that is centered on the add-on object but has a width and height 0.6 times the width and height of the person. The second step uses features from this box to make the final prediction $l_{id}$. The features for the second step are obtained by using ROI Pooling on $conv5$ (the last convolutional feature map), followed by two fully connected layers. The parameters of the fully connected layers are shared between the two steps. This helps the model zoom in to the neighborhood of the add-on object to make the final prediction.
\end{enumerate}

The CNN architecture we use is either Alexnet or VGG. The location predictions $l_{id}$ are obtained by a linear layer on top of the final fully connected layer from the second step. The confidences $p_{id}$ are obtained by a linear layer on top of the final fully connected layer from the first step, passed through a sigmoid.

We train this CNN on ground truth people. To do this, we map each person in the training set to all the add-on objects that are attached to her as follows: first, given an add-on object with segmentation mask $M$ and a person with segmentation mask $M'$, we measure the distance between them using the Hausdorff distance:
\begin{equation}
d(M, M') = \max_{x \in M} \min_{y \in M'} \|x-y\|_2
\end{equation}
We then assign each add-on object to the closest person in the image. If there is no person in the image, the add-on object is discarded. After this assignment, a person can end up being attached with multiple objects belonging to the same add-on category. To ensure that each person is attached to at most one object of each add-on category, we simply pick one of them and discard the rest.

This gives us a dataset of people annotated with attached objects. For each person $d$ for each add-on category $i$, we can now get a ground truth label $y_{id} \in \{0,1\}$ indicating whether there is an attached object of this category and a ground truth location $l^*_{id}$ if the attached category is present. We also construct the intermediate box target $\hat{l}^*_{id}$ with the same center as $l^*_{id}$ but a width and height 0.6 times that of the person. The network is then trained to minimize the loss
\begin{eqnarray}
\sum_{i,d} L_{\mathrm{nll}} (y_{id}, p_{id})  + y_{id} (L_{\mathrm{H}}(l^*_{id} - l_{id}) + L_{\mathrm{H}}(\hat{l}^*_{id} - \hat{l}_{id}))
\end{eqnarray}
where $L_{\mathrm{nll}}$ is negative log likelihood, and $L_{\mathrm{H}}$ is the Huber loss.

\begin{figure}
\centering
\includegraphics[height=0.08\textwidth]{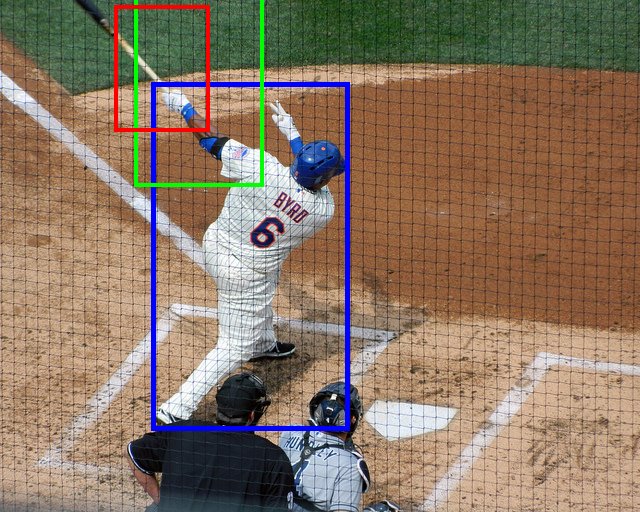}
\includegraphics[height=0.08\textwidth]{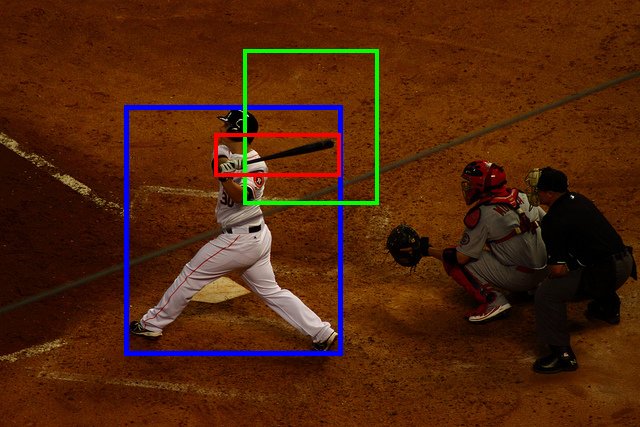}
\includegraphics[height=0.08\textwidth]{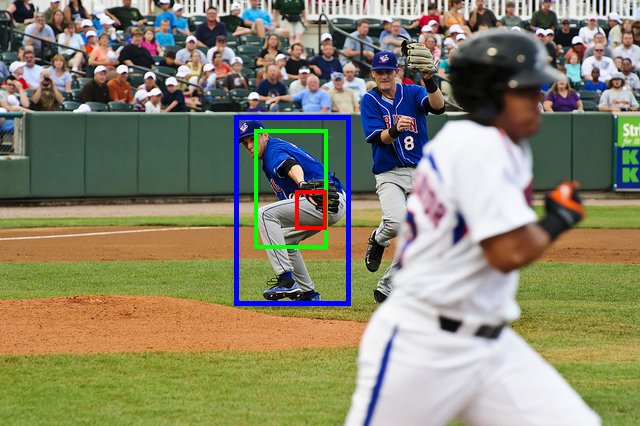}
\includegraphics[height=0.08\textwidth]{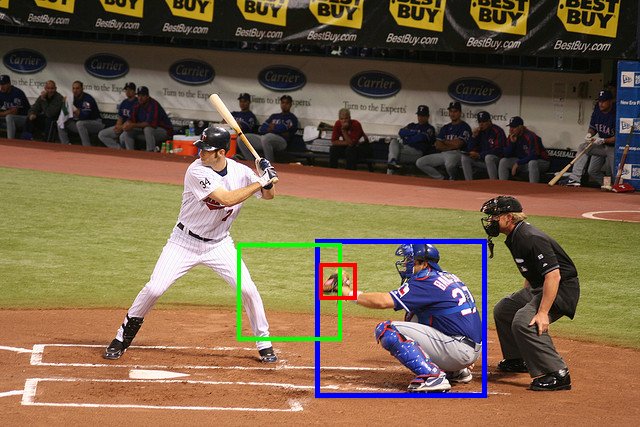}\\
\includegraphics[height=0.08\textwidth]{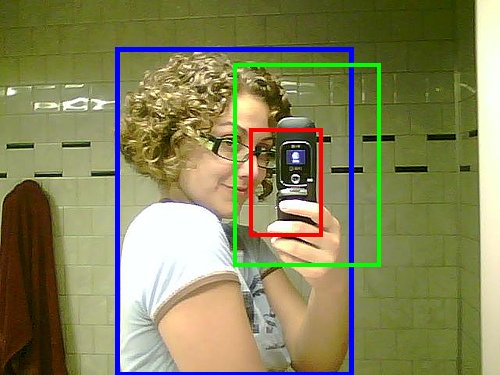}
\includegraphics[height=0.08\textwidth]{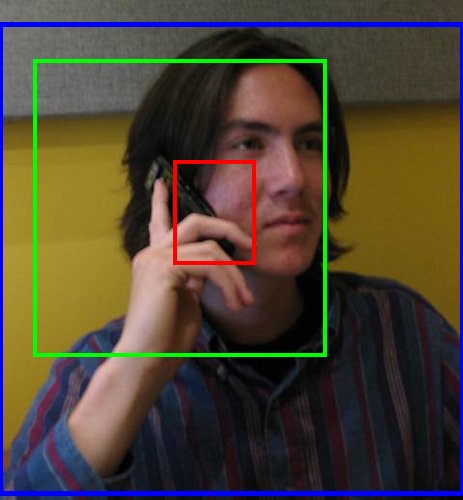}
\includegraphics[height=0.08\textwidth]{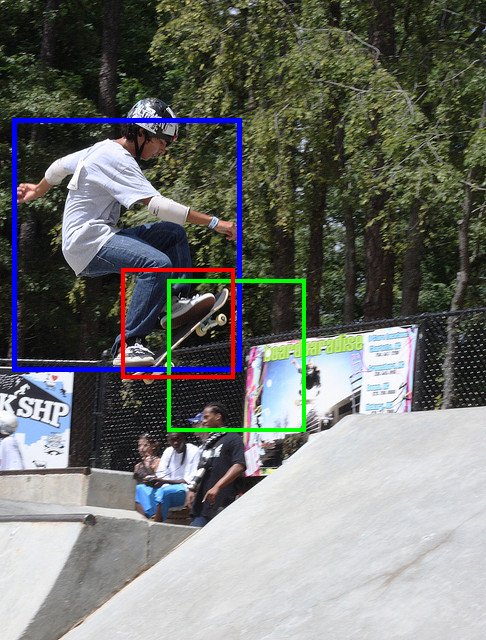}
\includegraphics[height=0.08\textwidth]{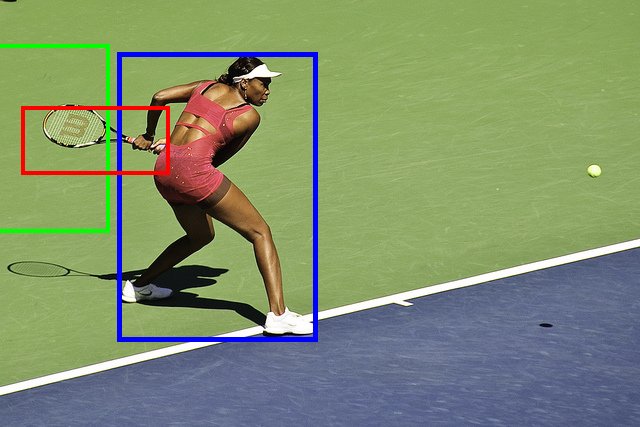}
\caption{Person-centric predictions for add-on categories: baseball bat, baseball bat, baseball glove, baseball glove, cell phone, cell phone, skateboard and tennis racket. In blue is the person, in green the first prediction and in red the second prediction.}
\label{fig:regresspreds}
\end{figure}

Figure~\ref{fig:regresspreds} shows the location predictions of person add-ons with the highest confidence. Quantitatively, when evaluated on the detection task, these predictions (computed on ground truth people boxes, trained with VGG) achieve an mAP of 30.5 at an overlap threshold of 0.1 and 6.1 at an overlap threshold of 0.5. For comparison, if we remove the iterated regression, mean AP at an overlap threshold of 0.1 drops to 16.3, indicating the importance of our iterative scheme. 
 While these numbers seem low, note that these predictions take into account a global view of the person, but they don't capture the add-on objects in detail. Next, we combine these person-centric predictions with local appearance features to get the final detections. 

%\todo{training details}
\subsection{Using person-centric predictions}
We get the person-centric predictions on Fast R-CNN person detections and convert them into an image-level heatmap (one for each add-on object category) as follows.  Given a person detection $d$ and the corresponding predicted bounding box $l_{id}$ and confidence $p_{id}$, we first take the center of $l_{id}$ and transform it to image coordinates to get $c_{id}$. We then construct a heatmap $s_{id}$ for this particular detection. The value of this heatmap at a pixel $p$ is:
\begin{equation}
s_{id}(p) =  A + \mathbb{I}[p_{id}>\theta]B\exp( - \frac{\|p-c_{id}\|^2}{2\sigma^2h_d^2})
\end{equation}
where $\mathbb{I}[\cdot]$ is 1 if its argument is true and 0 otherwise, and $h_d$ is the height of the person detection. Intuitively, if the confidence is greater than a threshold $\theta$, we paste a Gaussian centered at $c_{id}$ with a width proportional to $h_d$. We use $A=-50, B=100$ and $\sigma=0.001$.

We then produce a single heatmap for $i$  by taking the pointwise maximum over $d$:
\begin{equation}
s_i(p) = \max_d s_{id}(p)
\end{equation}

\begin{figure}
\includegraphics[width=0.45\textwidth]{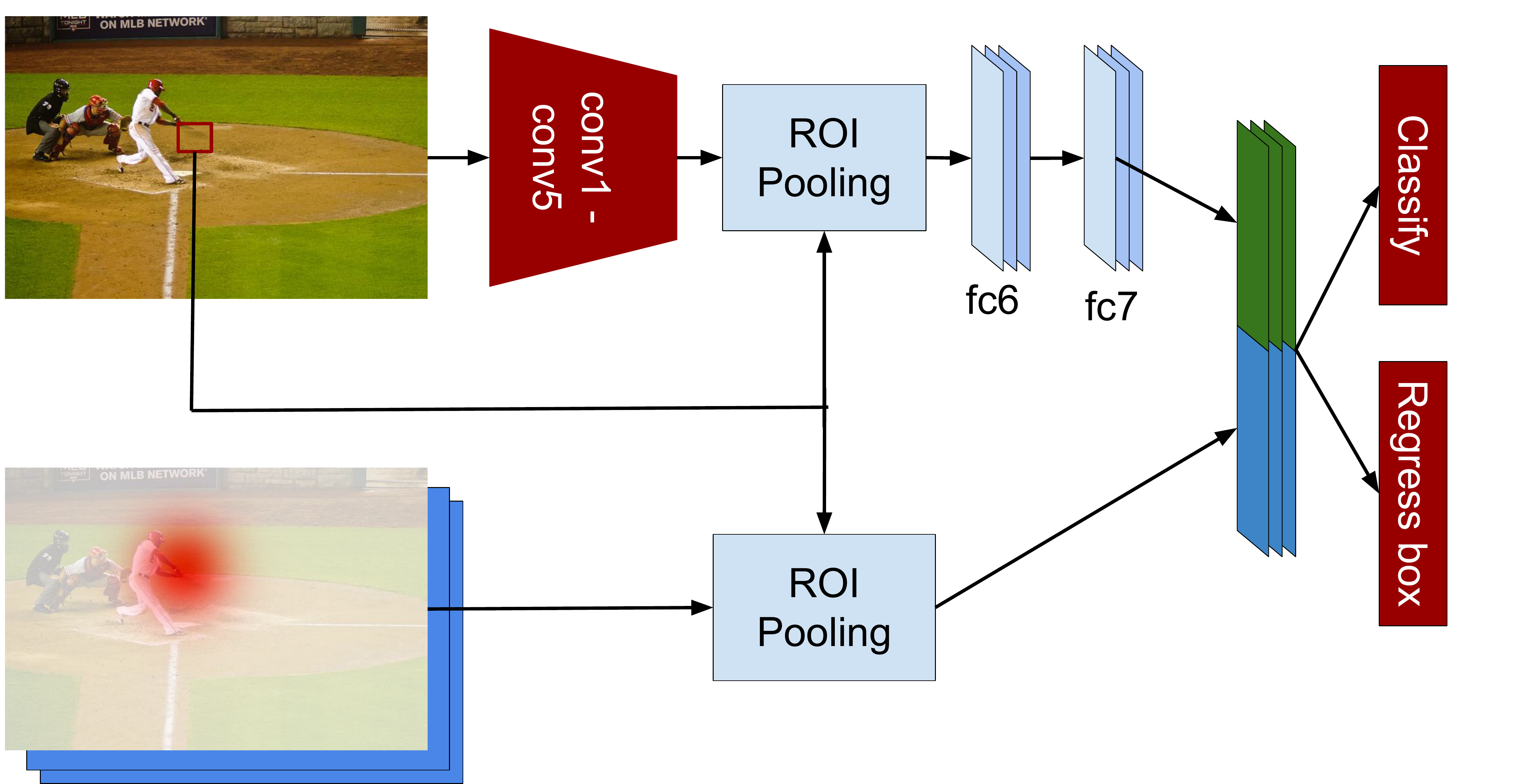}
\caption{Combining person-centric add-on predictions with local appearance features. The predictions are encoded as heatmaps and ROI Pooling is used to extract fixed-length feature vectors from this heatmap for each candidate bounding box. These features are then concatenated with the fc7 features produced by Fast RCNN and fed into the classifier and bounding box regressors.}
\label{fig:heatmap}
\end{figure}

This set of heatmaps forms a feature map with as many channels as there are add-on categories. For every box proposal,  we use ROI Pooling on this feature map to get a fixed length feature vector, which is then concatenated with the features from the last fully-connected layer of Fast RCNN and passed into the linear classifier and bounding box regressor. This entire architecture (shown in Figure~\ref{fig:heatmap} ) is trained end-to-end for object detection as in the standard Fast RCNN pipeline.

\subsection{Experimental results}
In Table~\ref{tab:onlyattached} we show final detection results on val01. We show the AP for all the add-on categories for both Alexnet and VGG. For both networks, almost all add-on categories improve, validating our intuition that reasoning about the person appearance can provide significant gains in detecting add-ons. For some categories the gain is especially large: in particular the AP improves by more than 6 points for both baseball bat and baseball glove. This might be because the interaction between people and these objects is especially well defined, and because the local appearance of these categories is not discriminative enough to do the job on its own.

Note that our model can only help improve detection performance for objects that are attached to people. The last four rows in Table~\ref{tab:onlyattached} show the mean AP when restricted to add-on objects that are attached to people (unattached add-on objects are marked with an ignore label: detections overlapping with them count as neither positives nor negatives). To get attached objects, we use the heuristic described in Section~\ref{subsec:personzoom}. We observe that the improvement offered by person context is larger in this setting, as expected. We get a 2.6 point improvement for Alexnet and a 1.9 point improvement for VGG. Note that the numbers are overall a bit lower in this setting, because these attached objects are in fact harder to detect owing to occlusion and small size.

\begin{figure}
\centering
\includegraphics[height=0.08\textwidth]{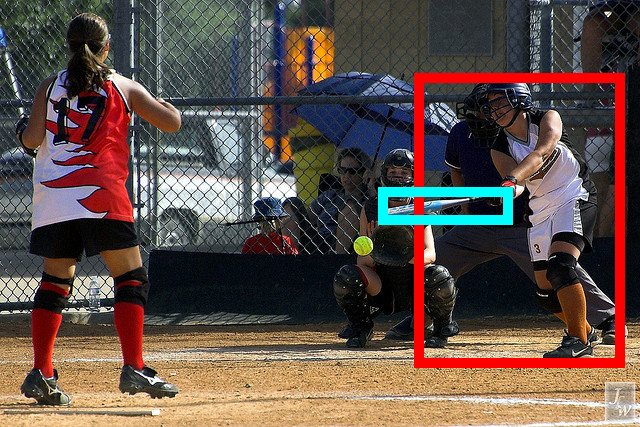}
\includegraphics[height=0.08\textwidth]{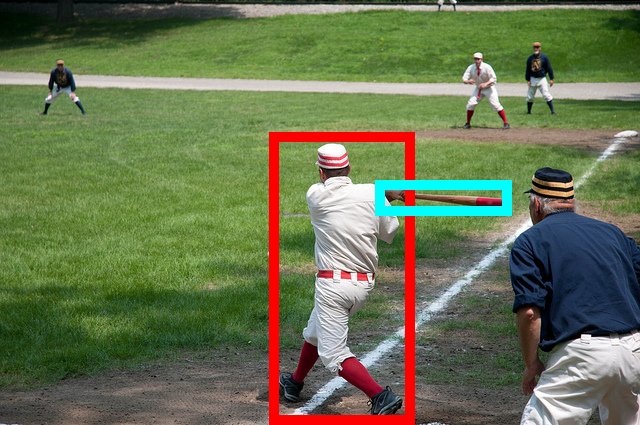}
\includegraphics[height=0.08\textwidth]{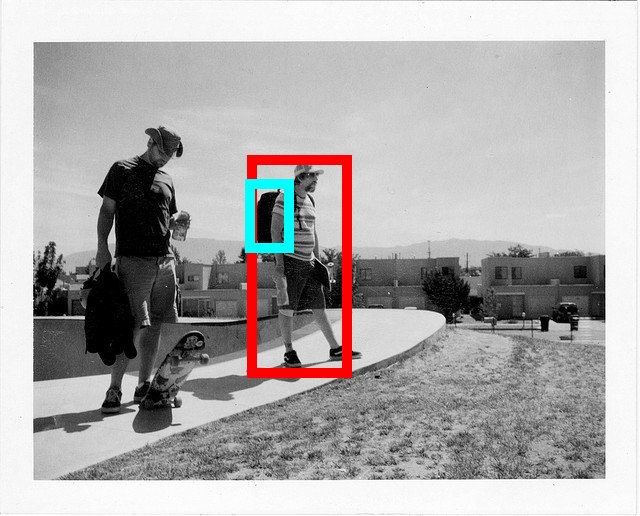}
\includegraphics[height=0.08\textwidth]{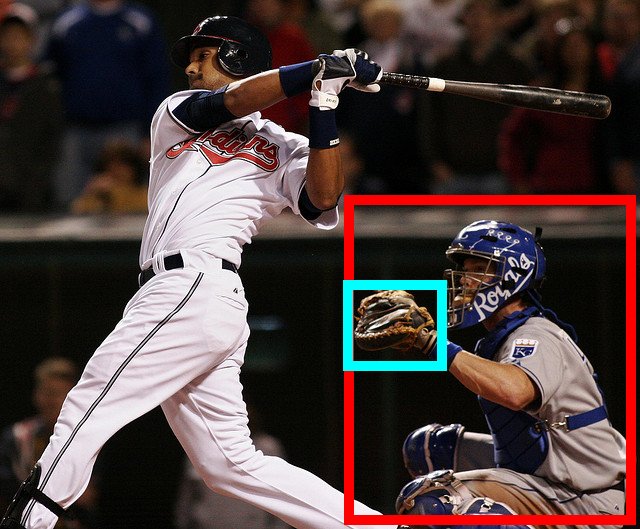}\\
\includegraphics[height=0.08\textwidth]{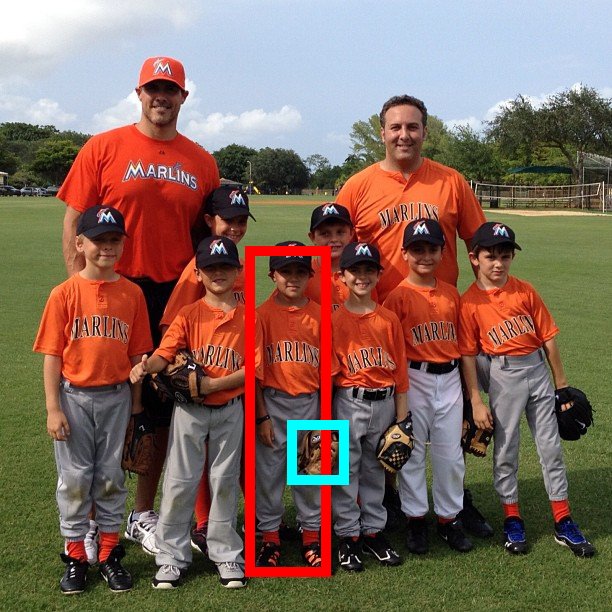}
\includegraphics[height=0.08\textwidth]{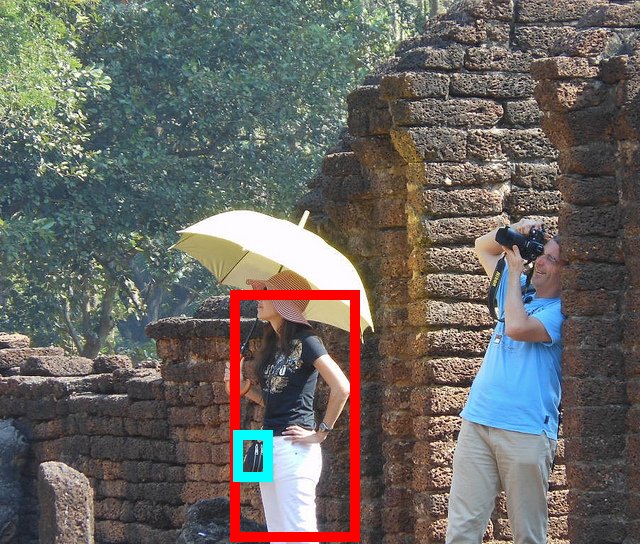}
\includegraphics[height=0.08\textwidth]{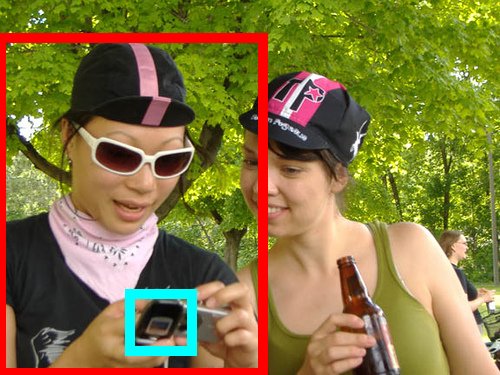}
\includegraphics[height=0.08\textwidth]{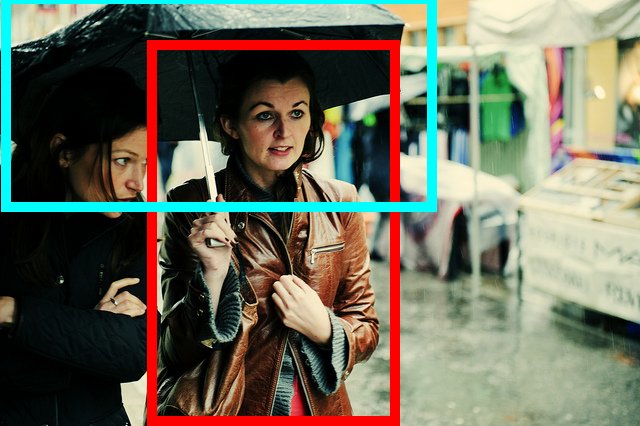}
\caption{Detections with inferred attachments: in order: baseball bat, baseball bat, backpack, baseball glove, baseball glove, handbag, cell phone, umbrella. The add-on objects are in blue and the people they are attached to are in red.}
\label{fig:attach}
\end{figure}

Our model also allows us to infer which add-ons are attached to which people. To do this inference, we take the center $c$ of each detection belonging to add-on category $i$, and take the person detection $d$ which gives the highest heatmap value $s_{id}(c)$ at that location for that category. Figure~\ref{fig:attach} shows the inferred attachments. The attachments make sense even in cluttered scenes, and allow us to detect objects that would otherwise be hard to detect.

\renewcommand{\arraystretch}{1.4} 
\setlength{\tabcolsep}{8pt}
\begin{table*}
\centering
\footnotesize
\resizebox{1.0\linewidth}{!}{
\begin{tabular}{lllcccccccccccccccg}
\toprule 
Method         & CNN     & Evaluation         & \vertical{back pack} & \vertical{umbrella} & \vertical{handbag} & \vertical{tie} & \vertical{suitcase} & \vertical{skis} & \vertical{snow board} & \vertical{sports ball} & \vertical{baseball bat} & \vertical{baseball glove} & \vertical{skate board} & \vertical{surf board} & \vertical{tennis racket} & \vertical{wine glass} & \vertical{cell phone} & \vertical{mean} \\
\midrule
Baseline       & Alexnet & all instances      & 7.0                   & 25.9                 & 4.0                 & \textbf{24.0}   & 20.1                 & 8.2              & 18.5                   & 24.0                    & 6.2                      & 20.0                       & 28.4                    & 23.2                   & 45.4                      & 17.4                   & 16.9                   & 19.3 \\
Person Context & Alexnet & all instances      & \textbf{7.3}          & \textbf{27.0}        & \textbf{4.2}        & 23.9            & \textbf{21.4}        & \textbf{8.9}     & \textbf{18.6}          & \textbf{26.3}           & \textbf{9.5}             & \textbf{27.1}              & \textbf{29.6}           & \textbf{25.2}          & \textbf{47.5}             & \textbf{17.4}          & \textbf{18.1}          & \textbf{20.8} \\
Baseline       & VGGnet  & all instances      & 15.7                  & 37.0                 & 10.5                & 32.3            & 35.0                 & 11.6             & 27.4                   & 37.2                    & 21.4                     & 37.5                       & 46.0                    & \textbf{38.7}          & \textbf{62.8}             & 28.6                   & \textbf{29.1}          & 31.4 \\
Person Context & VGGnet  & all instances      & \textbf{15.8}         & \textbf{37.3}        & \textbf{11.8}       & \textbf{34.0}   & \textbf{35.9}        & \textbf{12.5}    & \textbf{28.2}          & \textbf{39.2}           & \textbf{27.7}            & \textbf{43.2}              & \textbf{47.3}           & 38.4                   & 62.4                      & \textbf{29.0}          & \textbf{29.1}          & \textbf{32.8} \\
\midrule
Baseline       & Alexnet & attached instances & 5.2                   & 28.2                 & 3.2                 & 25.3            & 12.8                 & 8.4              & \textbf{18.2}          & 16.3                    & 6.8                      & 20.0                       & 28.8                    & 24.4                   & 45.5                      & 11.4                   & 14.9                   & 18.0 \\
Person Context & Alexnet & attached instances & \textbf{7.0}          & \textbf{29.9}        & \textbf{4.4}        & \textbf{26.9}   & \textbf{16.1}        & \textbf{9.8}     & 18.0                   & \textbf{20.4}           & \textbf{12.6}            & \textbf{28.2}              & \textbf{30.9}           & \textbf{27.1}          & \textbf{50.0}             & \textbf{11.9}          & \textbf{16.7}          & \textbf{20.6} \\
Baseline       & VGGnet  & attached instances & 12.3                  & 39.9                 & 9.3                 & 33.6            & 24.9                 & 11.5             & 28.1                   & 27.1                    & 23.4                     & 37.6                       & 46.3                    & \textbf{40.1}          & \textbf{62.9}             & 23.3                   & 24.6                   & 29.7 \\
Person Context & VGGnet  & attached instances & \textbf{12.8}         & \textbf{40.5}        & \textbf{10.7}       & \textbf{36.1}   & \textbf{26.4}        & \textbf{12.4}    & \textbf{29.0}          & \textbf{30.9}           & \textbf{31.1}            & \textbf{43.7}              & \textbf{47.6}           & 40.0                   & 62.4                      & \textbf{24.2}          & \textbf{25.4}          & \textbf{31.6} \\
\bottomrule
\end{tabular}}
\caption{Mean AP at 50\% overlap of the Person Context model and the Fast R-CNN Baseline on val01. Top four rows show performance on all instances and the bottom four show performance only on attached objects. See text for details.}
\label{tab:onlyattached}
\end{table*}

\section{Local Scene Context}
\label{sec:scenecontext}
The second model we propose in this paper captures more general contextual relationships between objects, and between objects and scenes. This generalizes person context, because now we are considering a larger variety of visual cues as context. It also means that the contextual relationships are not as precise: while a person will hold at most one baseball bat, and the baseball bat will be in his or her hands, a road can have many cars on it and they may be anywhere as long as they are on top of the road. A final challenge is that while in the previous case the contextual region was hand picked, now we do not \apriori know what the contextual region is and we need to pick it automatically. Because of these differences, the architecture we use for capturing this kind of context is different.

\subsection{Picking contextual regions}
We first need to decide which regions we are going to use as contextual regions. Intuitively, we should pick those regions that are informative of both object presence and location. However, finding such regions is a hard problem. Instead, we assume that if a region can confidently and accurately say that a particular object is \emph{present}, it can also tell us about the object's \emph{location}. Thus we are looking for image regions that are discriminative for the presence of particular object categories.

Such regions are discovered automatically by image classification algorithms based on Multiple Instance Learning (MIL). These algorithms treat each image $I$ as a bag $B$ of image regions: thus $B=\{ r: r \subset I\}$, where each region $r$ is a bounding box. Each region $r$ is scored with a classifier that assigns a probability $p_{ir}$ for each category $i$ indicating its confidence that an object of this category is present in the image. In the Noisy-Or model for image classification~\cite{Viola05}, these probabilities are combined to give a probability to the entire image:
\begin{equation}
p_{i} = 1 - \prod_r (1-p_{ir})
\label{eq:noisy-or}
\end{equation}
Intuitively, $p_i$ will be high if at least one of the probabilities $p_{ir}$ is close to 1. For training, each image is labeled with the set of categories present in the image, and the model is trained by minimizing the sum of negative log likelihoods for each category i.e., minimizing $\sum_i L_{nll} (y_i, p_i)$ where $y_i$ is 1 if the image in question contains an object of category $i$ and 0 otherwise.

We train such a model for image classification by adapting the Fast R-CNN architecture to minimize the loss described above instead. In particular, features for each region from the final fully connected layer are fed into a set of linear classifiers, one per object category, followed by sigmoids to produce the probabilities $p_{ir}$. These probabilities are then combined using Equation~\ref{eq:noisy-or} and fed into the loss. Our VGG MIL model achieves a classification mean AP of 76.9\% (77.0\% for person add-ons), indicating that predicting the presence of object categories works quite well. Note that layers conv1 through fc7 for this network were initialized using weights obtained by training Fast R-CNN out-of-the-box on COCO (as described in Section \ref{sec:prelims}).

We use this model to heuristically pick context regions as follows. For each region we compute the maximum probability it assigns to any class: $p_r = \max_i p_{ir}$. A high $p_r$ means that the region $r$ is confident about the presence of at least one object category in the image. We sort the regions in decreasing order of $p_r$ and pick the top $T=15$ regions for our context. To capture global context, we also add into this pool the region corresponding to the full image.

\subsection{Using context to score candidate boxes}
Let us first focus on a single context box $c$, and consider how we may use this context box to classify other candidate boxes in the image. Given any other box $b$ in the image, we want to produce a feature vector $\phi(c,b)$ that encapsulates what context $c$ has to say about box $b$. What $c$ says about $b$ depends on both the content of the context box as well as the spatial relationship between the two. We encode the spatial relationship between $c$ and $b$ using a set of binary valued functions $R_k(c,b)$, $k=1,\ldots, K$. Each function is 1 if $b$ is in a particular spatial relationship to $c$ and 0 otherwise. We consider two sets of these functions:
\begin{enumerate}
\item We take a large set of pairs of boxes and cluster the spatial relationships between them. Then we define a function $R^{cluster}_k$ for each cluster $k$ that outputs 1 if a pair belongs to that cluster and 0 otherwise. The representation we use to cluster the spatial relationships is as follows. Given two boxes $b$ and $c$, we compute the displacement $\delta$ between the two centers in units of the width and height of $c$. To prevent the values from growing very large, we transform $\delta$ as $\delta'=\mathrm{sign}(\delta)\log(1+\mathrm{abs}(\delta))$. We then concatenate with this the log of the width and height of $b$ in units of the width and height of $c$. We use k-means to get 48 clusters using this representation. 
\item We define an additional set of functions $R^{ov}_k$ where $R^{ov}_k$ is 1 if the two boxes overlap by more than or equal to a threshold $\theta_k$ and 0 otherwise. We use six thresholds: 0.5, 0.6, 0.7, 0.8, 0.9 and 1.
\end{enumerate}

\renewcommand{\arraystretch}{1.3} 
\setlength{\tabcolsep}{4pt}
\begin{table}
\centering
\footnotesize
\resizebox{1.0\linewidth}{!}{
\begin{tabular}{lccccgg}
\toprule
& \multicolumn{2}{c}{Alexnet on \textit{val01}} & \multicolumn{2}{c}{VGGnet on \textit{val01}} & \multicolumn{2}{g}{VGGnet on \textit{testdev}} \\
\cmidrule(lr){2-3} \cmidrule(rl){4-5} \cmidrule(lr){6-7}
Method                         & all           & person        & all                           & person               & all           & person \\
                               & objects       & add-ons       & objects                       & add-ons              & object        & add-ons \\
\midrule
FRCNN baseline                 & 30.8          & 19.3          & 41.9 (47.2)                   & 31.4 (40.4)          & 42.6          & 31.1\\
Global context                 & 31.4          & 20.0          &                               &                      &               & \\
Immediate neighborhood         & 31.3          & 20.4          &                               &                      &               & \\
Local Scene Context-Full       & \textbf{33.1} & \textbf{22.5} & \textbf{42.8} (\textbf{49.0}) & 32.7 (\textbf{43.0}) & \textbf{43.3} & 32.3\\
Local Scene Context-Coarse     & 32.5          & 21.3          &                               &                      &               & \\
Local Scene Context-Linear     & 32.5          & 21.7          &                               &                      &               & \\
Person Context                 & 31.0          & 20.8          & 42.2 (47.6)                   & \textbf{32.8} (42.4) & 42.9          & \textbf{32.6}\\
\bottomrule
\end{tabular}}
\caption{Mean AP at 50\% overlap on val01 and test-dev, of the Local Scene Context model, and several baselines and ablations. The numbers in parentheses show the mean AP when ground-truth bounding boxes are added to the pool of object proposals. The last column (in gray) shows performance on test-dev. Per-category numbers are shown in the supplementary material.}
\label{tab:localscenecontext}
\end{table}

This defines $K=54$ functions.
We then use the following form for $\phi(c,b)$:
\begin{eqnarray}
\phi(c,b) &=& \sum_{k=1}^K R_k(c,b) \phi_k(c) \\
\phi_k(c) &=& W_k f(c)
\end{eqnarray}
Here $f(c)$ is the feature vector for the context box, obtained from the last fully-connected layer in the Fast R-CNN network, and the $W_k$ are weights that are learned.

Since we are effectively adding $K$ different fully connected layers, to prevent the number of parameters from blowing up we keep the $\phi_k(c)$ low dimensional: in our experiments, they have a dimensionality of $d_1$=40.

Each box $b$ now gets a $d_1$-dimensional feature vector $\phi(c,b)$ from each contextual region $c\in \mathcal{C}$. Now we need to combine these feature vectors. However, this combination needs to be able to capture sophisticated reasoning. For instance, global scene features might tell us that we are looking at a frisbee game, and the pose of a player might tell us that the player is reaching for something, and we need to combine this information to find the frisbee. This suggests that we want to combine these feature vectors non-linearly. We adopt the strategy of transforming the feature vectors non-linearly into a high-dimensional space and accumulating evidence in that space. Concretely, we first linearly project all the feature vectors $\phi(c,b)$ to a high dimensional space of dimensionality $d_2$ = 2048. We then pass these high dimensional vectors $\psi(c,b)$ through a non-linearity $h_1(\cdot)$, sum them together, and pass the sum through another non-linearity $h_2$ to get the final context feature vector for this box $\psi_{context}(b)$. We experimented with two alternatives for $h_1$ and $h_2$: tanh and ReLU. The tanh nonlinearities performed better by 2.5 points mean AP on val01 using Alexnet. Our model thus uses tanh as the nonlinearity:

\begin{eqnarray}
\psi(c, b) &=& W^{(2)} \phi(c,b) \label{eq:comb1} \\
\psi_{context}(b) &=& \mathrm{tanh}( \sum_{c \in C} \mathrm{tanh}(\psi(c,b))) \label{eq:comb2}
\end{eqnarray}
Finally, these context features are concatenated with the appearance features of the box obtained from the final fully connected layer in the Fast R-CNN network and passed into a linear classifier and a bounding box regressor. The entire network is trained end-to-end for detection.
\subsection{Experiments}
In Table~\ref{tab:localscenecontext} we show the final detection results on val01. We show the mean AP on all categories and the mean AP on the add-on categories mentioned in Section~\ref{sec:personcontext}. Apart from Fast R-CNN, we compare to two other baselines. ``Global context" uses the same architecture as ours, but uses a single context region corresponding to the entire image. ``Immediate neighborhood" uses two regions, one twice the size of the object candidate and the other four times the size, and concatenates features from all three together. We improve significantly over all three baselines, getting a 2.5 point boost over Fast R-CNN for the Alexnet network. For VGG, the gains are smaller but still statistically significant (measured with a paired t-test, $p<0.01$). Compared to the Person Context model, this model does as well (for VGG) or better (for Alexnet) on the add-on categories and is better overall.   Note that the training procedure for Local Scene Context had several subtle differences with respect to Fast R-CNN training (explained in the supplementary material). A Fast R-CNN model trained with the same settings performs much worse (41.3 mean AP, 30.5 on person add-ons), indicating that there is room for improvement in the Local Scene Context model, and a different training setting may offer larger gains.

\renewcommand{\arraystretch}{1.2} 
\setlength{\tabcolsep}{8pt}
\begin{table}
\centering
\footnotesize
\resizebox{1.0\linewidth}{!}{
\begin{tabular}{lccccc}
\toprule
& \multicolumn{2}{c}{Alexnet on \textit{val01}} & & \multicolumn{2}{c}{VGGnet on \textit{val01}} \\
\cmidrule(r){2-3} \cmidrule(r){5-6} 
Method                                 & all          & person       &  & all           & person   \\
                                       & objects      & add-ons      &  & objects       & add-ons  \\
\midrule
FRCNN baseline                         & 6.4          & 8.0          &  & 14.0          & 16.2     \\
Local Scene Context-Full               & \textbf{8.5} & \textbf{10.4}&  & \textbf{15.7} & \textbf{18.0} \\
Person Context                         & 6.8          & 9.2          &  & 14.3          & 17.2        \\
\bottomrule
\end{tabular}}
\caption{Mean AP at 50\% overlap of the Local Scene Context model, the Person Context model, and the Fast R-CNN baseline on small objects on val01.}
\label{tab:small}
\end{table}

We also show the results of two ablated versions of our system. ``Local Scene Context - Coarse" uses only 4 clusters of spatial relationships instead of 54.  This model doesn't perform as well, suggesting that capturing precise spatial relationships is necessary. ``Local Scene Context - Linear" removes the non-linear feature combination (Equations~\ref{eq:comb1} and \ref{eq:comb2}) and simply adds the $\phi(c,b)$ together. This works much worse, confirming that a linear combination rule isn't sufficient to capture contextual information faithfully.

One reason for why the gains are not so high is that there is an implicit upper bound because of the region proposals: classifying region proposals better won't help unless the right region proposals are actually there in the pool. This is especially true for our model (as opposed to the Fast RCNN baseline) because the categories where context can help is precisely the kind of small, hard to detect objects for which region proposals aren't good enough. To test this hypothesis, we added ground truth boxes into the candidate pool at test time, and computed the performance of all three methods : Fast R-CNN, Person Context and Local Scene Context. The numbers are shown in parentheses in Table~\ref{tab:localscenecontext}. While all three models gain significantly in this setting (as expected), the gap between the contextual models and the baseline also increases. In particular, for the Local Scene Context models, the improvement goes up from about 0.9 points to 1.8 points. This suggests that the paucity of good proposals may be the bottleneck in performance, and the gains from reasoning about context will be more prominent as proposal methods become better.

\begin{figure}
\centering
\includegraphics[height=0.075\textwidth]{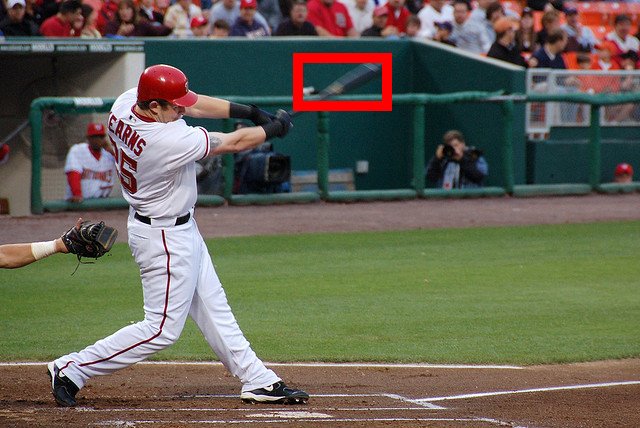}
\includegraphics[height=0.075\textwidth]{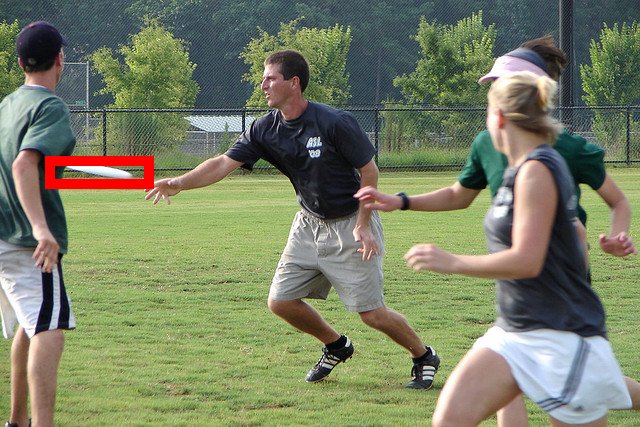}
\includegraphics[height=0.075\textwidth]{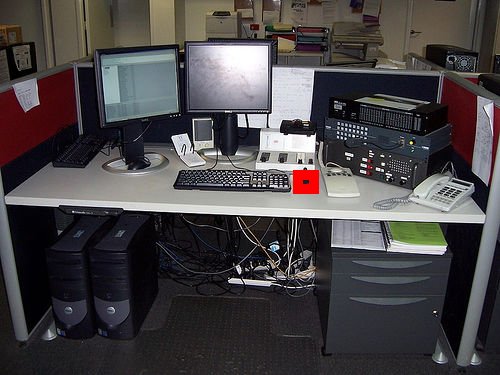}
\includegraphics[height=0.075\textwidth]{figs_jpg/soth/sportsball_1_2479_0.jpg}\\
\includegraphics[height=0.075\textwidth]{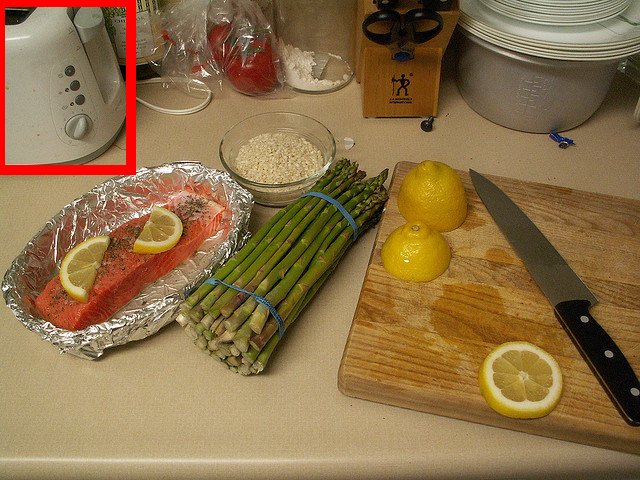}
\includegraphics[height=0.075\textwidth]{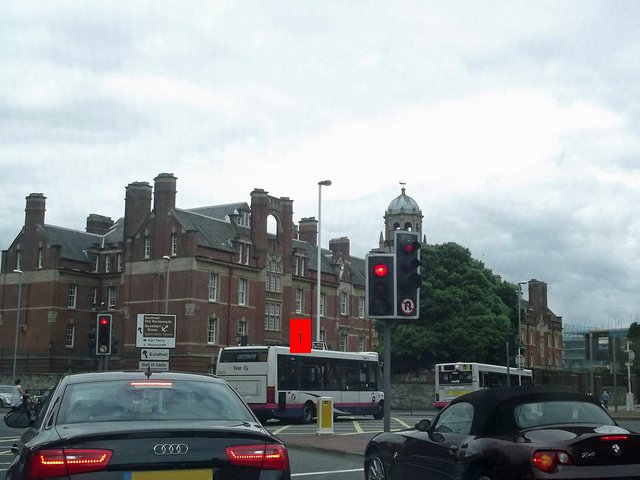}
\includegraphics[height=0.075\textwidth]{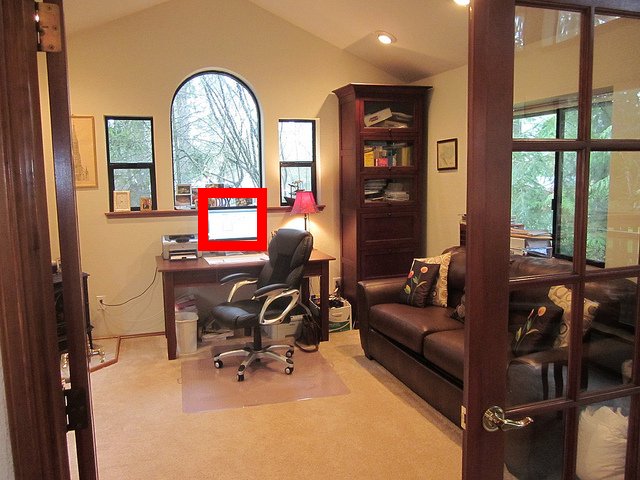}
\includegraphics[height=0.075\textwidth]{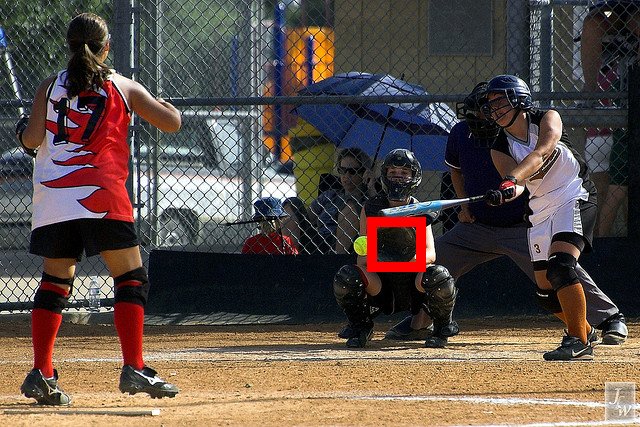}
\caption{High scoring detections from the Local Scene Context model that are missed by the baseline Fast RCNN model. In order: baseball-bat, frisbee, mouse, sports-ball, toaster, traffic-light, tv, baseball-glove. Context helps a lot for small (the traffic light and the sports ball) or heavily occluded (the toaster) objects.}
\label{fig:scenecontextdets}
\end{figure}

\begin{figure}
\centering
\includegraphics[width=0.45\textwidth]{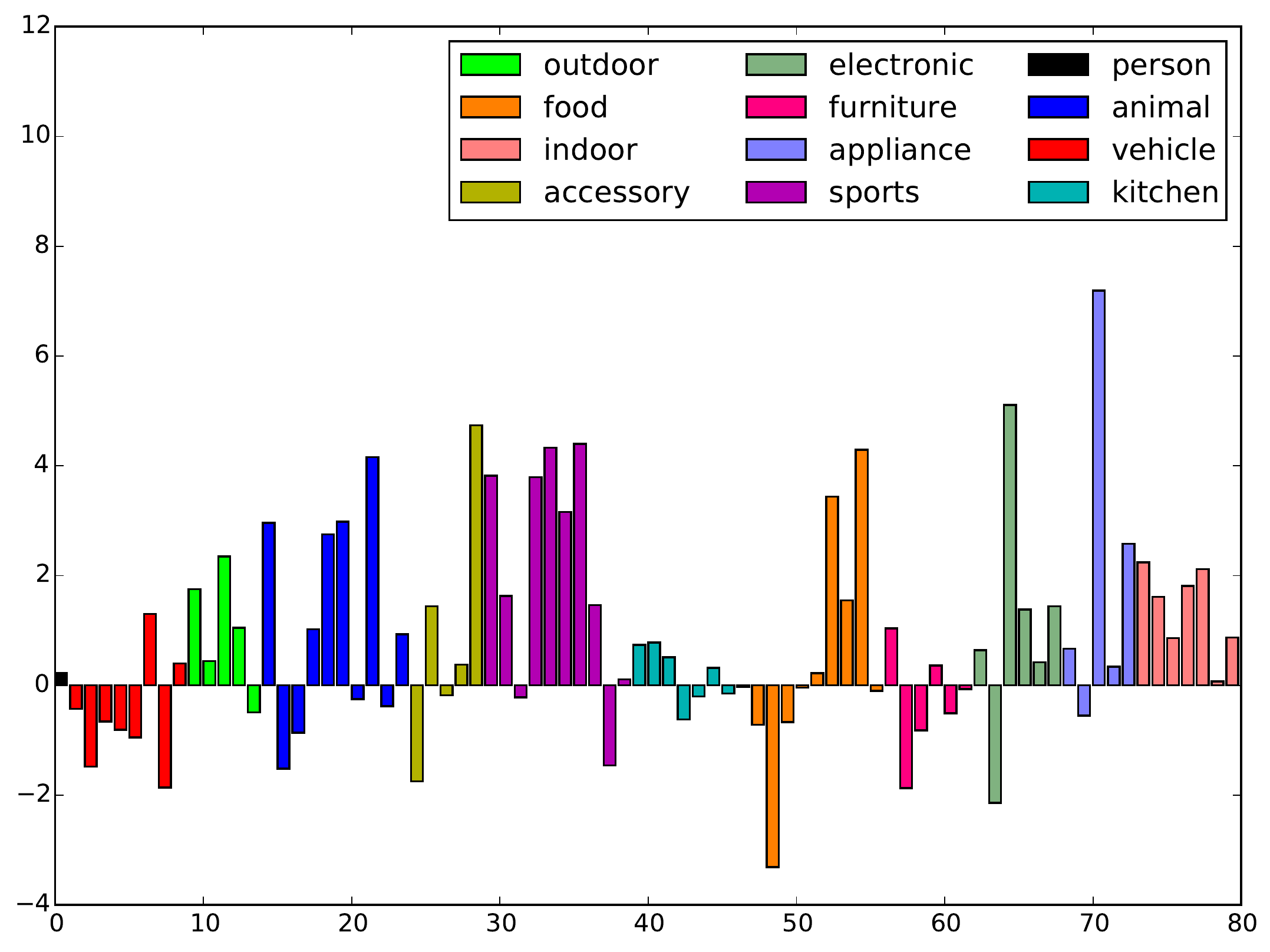}
\caption{Absolute improvements in AP over the baseline, color-coded by the COCO supercategory label.}
\label{fig:colorcodeexp}
\end{figure}

An improvement in mean AP is not very informative: a small improvement might be spread over a large number of categories, or might reflect a large gain in a single class. While it is hard to look at every class individually, we can divide the set of classes into a few supercategories and look at the distribution of performance in each. Supercategory labels are provided by COCO. In Figure~\ref{fig:colorcodeexp} we plot the improvements in each category (AP for Local Scene Context-Full minus AP for Fast RCNN), color coded by the super-category to which it belongs. We see that some super-categories improve almost across the board: ``outdoor" (traffic-lights, stop-signs, fire hydrants etc.), ``sports" (baseball bats and gloves, tennis rackets etc.), ``electronic" (tv, laptop, mouse etc.), ``appliance" (oven, toaster, sink etc.) and ``indoor" (book, clock, vase, toothbrush etc.). This makes sense, since these super-categories involve objects that occur in very prototypical contextual settings. Others such as ``vehicles" (car, aeroplane etc.) and ``furniture" (chair, couch, table etc.) seem to uniformly lose, although the losses are quite small. An interesting outlier is the ``animal" supercategory, which sees uniform gains. 

Another way to break up performance is in terms of the sizes of the objects we are trying to detect. Intuitively, context should be more useful for small objects where the visual signal itself may not be enough. Table~\ref{tab:small} shows the performance of all three models (the Fast RCNN baseline, Local Scene Context and Person Context) on small objects (less than $32 \times 32$ pixels). The performance of all three models is much lower, but the local scene context model performs 1.7 points better than the baseline for VGG (10\% relative improvement) and 2 points better for Alexnet (32\% relative improvement). This large relative gain reaffirms that small objects are indeed where context helps the most.

\begin{figure}
\centering
\includegraphics[width=0.50\textwidth]{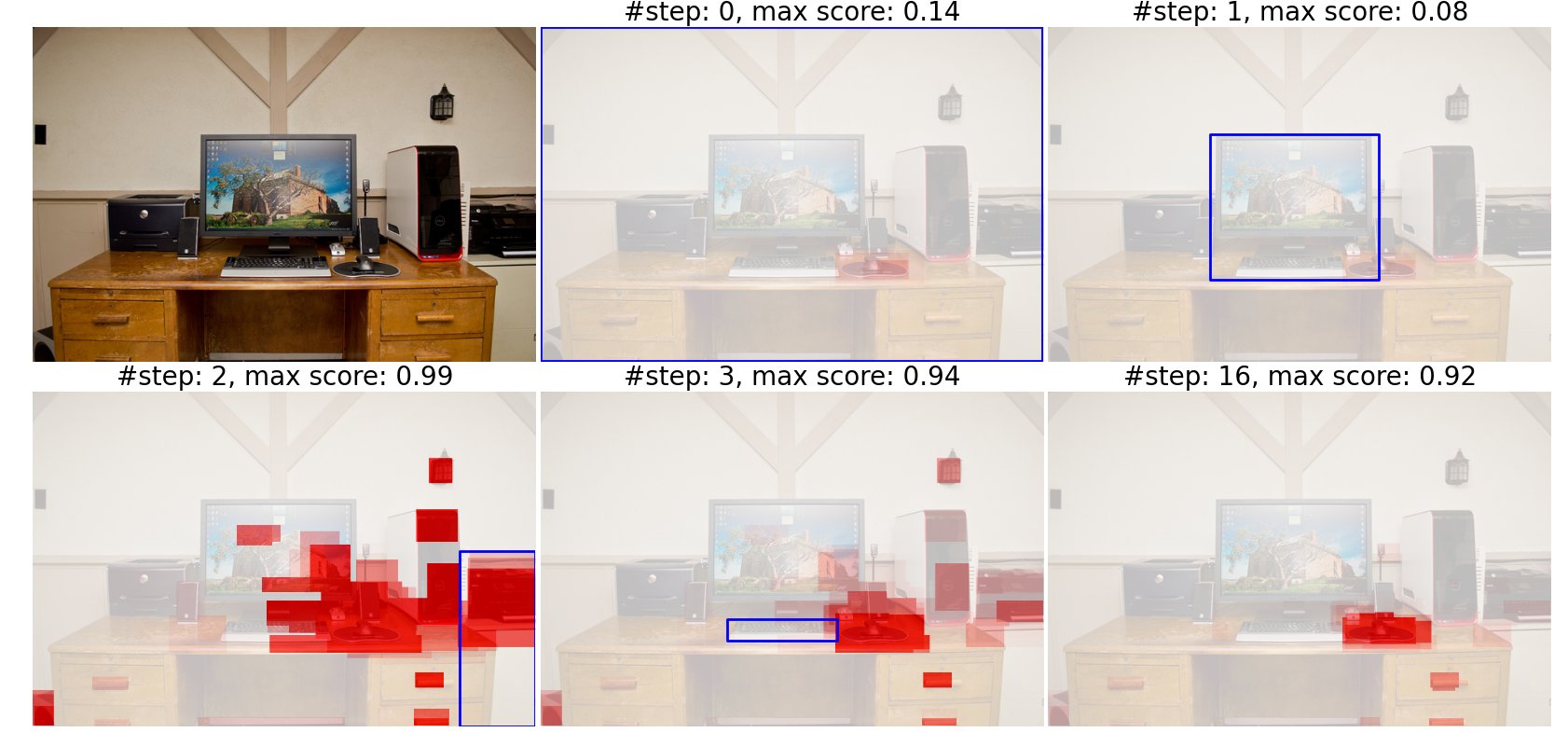}\\
\centering
\includegraphics[width=0.50\textwidth]{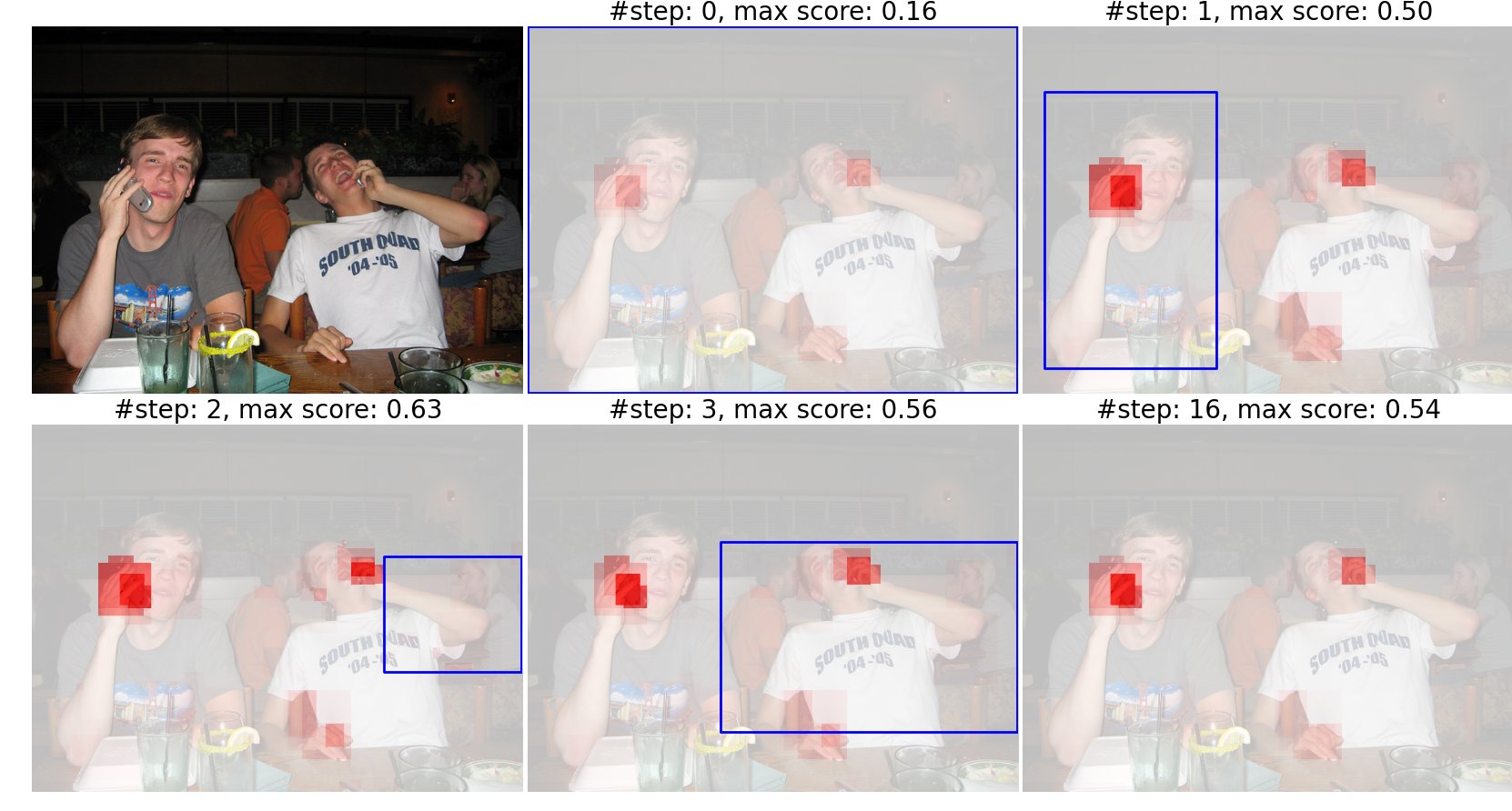}
\caption{Visualizing how context regions (in blue) change detection score. Red regions correspond to boxes that score high. First set: cell phone, second set: mouse.}
\label{fig:scenecontextvis}
\end{figure}

Figure~\ref{fig:scenecontextdets} visualizes some detections from our model corresponding to objects that are missed by the Fast R-CNN baseline. As expected, these tend to be small or heavily occluded objects which are hard to detect except in context. Figure~\ref{fig:scenecontextvis} visualizes how the score of different boxes change as context regions are incorporated. This is hard to visualize in our full model owing to the tanh non-linearities, so we use the Local Scene Context - Linear model instead. Each ``step" in the visualization adds a context box to the scoring. To keep the scale of the context features meaningful, at each step $t$, we scale up the context features by $16/t$. For $t=0$, we visualize the scores without the context features. 
Finally, the last column of Table~\ref{tab:localscenecontext} shows results on the test-dev set. As on the validation set, we get small but statistically significant improvements (paired t-test, $p<0.01$).

\section{Conclusion}
In this paper we have proposed two context models: one specific to people and their fashion and sports accessories, and the other capturing general relationships among objects and between objects and scenes. We have shown that these ideas provide significant gains even in the era of high-powered object detectors using CNNs, especially for small, occluded and hard-to-detect objects. 

\paragraph{Acknowledgments: } This work was supported by ONR SMARTS MURI
N00014-09-1-1051, a Berkeley Graduate Fellowship, and a Google Fellowship in
Computer Vision. We gratefully acknowledge NVIDIA corporation for the Tesla and
Titan GPUs used for this research.

{\small
\bibliographystyle{ieee}
\bibliography{refs}
}
\balance
\clearpage

\section*{Supplementary Material}
\newcommand{\tif}[1]{{\color{red}#1}}
\newcommand{\tbf}[1]{{\color{blue}#1}}

\definecolor{Gray}{gray}{0.85}
\newcolumntype{g}{>{\columncolor{Gray}}c}

\definecolor{Green}{rgb}{0.6,1.0,0.6}
\newcolumntype{o}{>{\columncolor{Green}}c}

\definecolor{Yellow}{rgb}{1.0,1.0,0.6}
\newcolumntype{y}{>{\columncolor{Yellow}}c}

\newcommand{\megaTable}[3]{
\renewcommand{\arraystretch}{1.4} 
\setlength{\tabcolsep}{8pt}
\begin{table*} \centering \footnotesize 
\resizebox{1.0\linewidth}{!}{\begin{tabular}{llyyyyyyycccoooggg}
\toprule 
& \verticala{Super Category} & 
\verticala{FRCNN baseline} & \verticala{Global context} & \verticala{Immediate
neighborhood} & \verticala{Local Scene Context-Full} & \verticala{Local Scene
Context-Coarse} & \verticala{Local Scene Context-Linear} & \verticala{Person
Context} & \verticala{FRCNN baseline} & \verticala{Local Scene Context-Full} &
\verticala{Person Context} &
\verticala{FRCNN baseline} & \verticala{Local Scene Context-Full} & \verticala{Person Context} &
\verticala{FRCNN baseline} & \verticala{Local Scene Context-Full} & \verticala{Person Context} \\ \midrule
% & & \multicolumn{13}{c}{val01} & \multicolumn{3}{g}{test-dev} \\ \midrule %cmidrule(lr){3-15} \cmidrule(lr){16-18}
& & \multicolumn{7}{y}{Alexnet on \textit{val01}} & \multicolumn{3}{c}{VGG on \textit{val01}} & \multicolumn{3}{o}{VGG on \textit{val01}} & \multicolumn{3}{g}{VGG on \textit{testdev}} 
\\ % \midrule %\cmidrule(lr){3-9} \cmidrule(lr){10-12} \cmidrule(lr){13-15} \cmidrule(lr){16-18}
& & & & & & & & & & & & \multicolumn{3}{o}{+GT Region} & & &\\ \midrule %\cmidrule(lr){13-15}
#1 \bottomrule 
\end{tabular}} 
\caption{#2} 
\tablelabel{#3}
\end{table*}}

\newcommand{\megaTableCaption}[0]{Average Precision at 50\% overlap for all
classes. We report performance for all methods reported in Table 2 in the main
paper. Absolute improvements of more than 1\% AP over the corresponding \fastrcnn
baseline are reported in \tbf{blue} (analogously absolute drops of more
than 1\% as compared to \fastrcnn is reported in \tif{red}).}

\megaTable{
                  & all            & 30.7 & 31.4       & 31.3       & \tbf{33.1} & \tbf{32.5} & \tbf{32.5} & 31.0       & 41.9 & 42.8       & 42.2       & 47.2 & \tbf{49}   & 47.6       & 42.6 & 43.3       & 42.9 \\
~~~~~~~~~~~~~~~~~~& person add-ons & 19.3 & 20.0       & \tbf{20.4} & \tbf{22.5} & \tbf{21.3} & \tbf{21.7} & \tbf{20.8} & 31.4 & \tbf{32.7} & \tbf{32.8} & 40.4 & \tbf{43}   & \tbf{42.4} & 31.1 & \tbf{32.3} & \tbf{32.6} \\ \\
                  & person         & 50.0 & 50.1       & \tbf{51.2} & \tbf{51.8} & 50.7       & 50.8       & 49.4       & 60.0 & 60.2       & 60.1       & 75.2 & \tbf{76.4} & 75.3       & 59.8 & 60.0       & 59.9 \\
                  & vehicle        & 40.6 & 40.2       & 41.1       & 41.5       & 41.2       & 41.4       & 40.6       & 51.6 & 51.0       & 51.5       & 56.6 & 56.5       & 56.4       & 52.1 & 52.1       & 52.3 \\
                  & outdoor        & 35.6 & 36.2       & 36.3       & \tbf{37.3} & \tbf{37.4} & \tbf{37}   & 35.5       & 45.0 & \tbf{46}   & 45.1       & 49.6 & \tbf{51.1} & 49.3       & 50.1 & 50.7       & 50.2 \\
                  & animal         & 55.1 & 55.7       & 54.7       & \tbf{57.4} & \tbf{57.2} & \tbf{56.8} & 55.2       & 67.9 & \tbf{69.1} & 68.0       & 71.3 & \tbf{73.4} & 71.4       & 67.5 & 68.5       & 67.5 \\
                  & accessory      & 16.2 & 16.3       & 16.8       & \tbf{18.3} & \tbf{17.7} & \tbf{17.7} & 16.7       & 26.1 & 27.0       & 27.0       & 30.2 & \tbf{31.8} & \tbf{31.4} & 27.0 & 27.8       & 27.3 \\
                  & sports         & 22.7 & \tbf{25.3} & \tbf{24.5} & \tbf{27.7} & \tbf{26.6} & \tbf{26.8} & \tbf{24.5} & 36.8 & \tbf{38.9} & \tbf{38.4} & 49.8 & \tbf{53.7} & \tbf{52.2} & 37.4 & \tbf{39.8} & \tbf{39.5} \\
                  & kitchen        & 13.8 & 13.5       & 14.1       & 14.6       & 14.1       & 13.9       & 14.0       & 23.7 & 23.9       & 23.9       & 31.1 & \tbf{32.3} & 31.2       & 23.3 & 23.1       & 23.3 \\
                  & food           & 26.7 & 26.3       & 26.8       & \tbf{28.3} & 27.7       & 27.6       & 26.9       & 36.2 & 36.7       & 36.1       & 39.0 & \tbf{40.5} & 38.9       & 34.8 & 35.2       & 34.9 \\
                  & furniture      & 36.2 & 36.5       & 36.7       & \tbf{38.7} & \tbf{37.3} & \tbf{37.6} & 36.2       & 47.1 & 46.8       & 47.1       & 50.1 & 50.5       & 50.1       & 45.6 & 45.6       & 45.7 \\
                  & electronic     & 31.5 & \tbf{33.2} & \tbf{33.2} & \tbf{35.3} & \tbf{34.2} & \tbf{34.8} & 31.7       & 45.6 & \tbf{46.7} & 45.6       & 48.4 & \tbf{50.3} & 48.7       & 47.4 & \tbf{49.1} & 47.3 \\
                  & appliance      & 29.4 & \tbf{31.9} & 29.2       & \tbf{32.8} & \tbf{32.8} & \tbf{32.3} & 29.7       & 42.0 & \tbf{44.1} & 42.0       & 44.0 & \tbf{46.4} & 43.9       & 44.3 & \tbf{46.1} & 44.3 \\
                  & indoor         & 18.5 & 19.0       & 19.3       & \tbf{20.3} & \tbf{19.8} & \tbf{19.9} & 18.1       & 26.6 & \tbf{27.9} & 26.7       & 31.3 & \tbf{33.8} & 31.6       & 29.3 & 28.7       & 29.4 \\
}{Mean Average Precision at 50\% overlap over all classes, person add-ons and
supercategories as defined in the COCO dataset. We report performance for all
methods reported in Table 2 in the main paper. Absolute improvements of more
than 1\% mean AP over the corresponding \fastrcnn baseline are reported in
\tbf{blue} (analogously drops of more than 1\% as compared to \fastrcnn are
reported in \tif{red}).}{summary}

\megaTable{
person         & person    & 50.0 & 50.1       & \tbf{51.2} & \tbf{51.8} & 50.7       & 50.8       & 49.4       & 60.0 & 60.2       & 60.1       & 75.2 & \tbf{76.4} & 75.3       & 59.8 & 60.0       & 59.9 \\ \\
bicycle        & vehicle   & 26.4 & 25.5       & 26.4       & 26.2       & 26.8       & \tbf{27.7} & 26.7       & 37.7 & 37.3       & \tif{36.6} & 43.5 & 43.1       & \tif{42.4} & 34.8 & 35.0       & 35.5 \\
car            & vehicle   & 27.4 & 26.6       & \tbf{29.4} & 27.1       & 26.7       & \tif{25.5} & 27.3       & 40.4 & \tif{38.9} & 40.5       & 53.0 & \tif{52}   & 53.0       & 39.8 & 38.8       & 40.1 \\
motorcycle     & vehicle   & 43.0 & 43.2       & \tbf{44.1} & \tbf{45.7} & 43.4       & \tbf{44.3} & 43.3       & 54.4 & 53.7       & 54.8       & 58.6 & 58.7       & 58.7       & 54.1 & 53.7       & 54.4 \\
airplane       & vehicle   & 53.9 & 54.3       & \tif{52.2} & \tbf{55.7} & 54.0       & \tbf{56}   & \tif{52}   & 63.9 & 63.1       & 64.1       & 69.0 & 68.4       & 68.7       & 67.0 & 67.6       & 66.4 \\
bus            & vehicle   & 54.8 & 54.1       & 54.7       & 55.7       & 55.7       & \tbf{56.6} & 55.2       & 68.1 & 67.1       & 67.5       & 68.4 & 68.3       & 68.1       & 74.3 & 75.3       & 74.4 \\
train          & vehicle   & 70.6 & \tif{69.1} & \tif{68.2} & 70.7       & 70.2       & 70.3       & 70.1       & 76.0 & \tbf{77.3} & 75.9       & 77.6 & \tbf{79.2} & 77.7       & 77.0 & 78.0       & 77.2 \\
truck          & vehicle   & 27.9 & \tif{26.4} & \tbf{30.4} & 27.1       & 27.6       & 27.5       & 28.7       & 41.4 & \tif{39.6} & 41.4       & 43.4 & \tif{41.8} & 43.4       & 40.2 & \tif{38}   & 40.8 \\
boat           & vehicle   & 20.8 & \tbf{22.5} & \tbf{23.2} & \tbf{23.8} & \tbf{25}   & \tbf{23.3} & 21.7       & 31.0 & 31.4       & 30.9       & 39.5 & \tbf{40.8} & 39.2       & 29.7 & 30.4       & 29.7 \\ \\
traffic light  & outdoor   & 17.9 & 18.0       & 18.8       & \tbf{20.1} & \tbf{19.3} & \tbf{19.3} & 17.7       & 29.0 & \tbf{30.8} & 29.5       & 42.8 & \tbf{46.3} & 42.0       & 29.9 & \tbf{32}   & 29.9 \\
fire hydrant   & outdoor   & 53.8 & 54.3       & 54.4       & \tbf{56.5} & \tbf{57}   & \tbf{55.3} & 53.7       & 68.7 & 69.2       & 68.5       & 72.5 & 72.8       & \tif{71.4} & 71.6 & 71.0       & 71.6 \\
stop sign      & outdoor   & 61.3 & \tbf{62.7} & 61.0       & \tbf{63.5} & \tbf{63.5} & \tbf{63.6} & 61.3       & 67.7 & \tbf{70}   & 68.2       & 69.7 & \tbf{71.7} & 70.3       & 69.4 & 69.3       & 69.5 \\
parking meter  & outdoor   & 28.7 & \tbf{30.1} & \tbf{30.9} & \tbf{29.9} & \tbf{30.8} & \tbf{30.3} & 28.5       & 35.1 & \tbf{36.1} & 35.4       & 36.8 & \tbf{38.1} & 37.1       & 52.8 & \tbf{54.4} & 53.1 \\
bench          & outdoor   & 16.5 & 15.8       & 16.3       & 16.4       & 16.6       & 16.7       & 16.4       & 24.4 & 23.9       & 24.0       & 26.2 & 26.7       & 25.8       & 26.6 & 27.0       & 27.1 \\ \\
bird           & animal    & 20.0 & \tbf{21.5} & 19.9       & \tbf{22.2} & \tbf{22.5} & \tbf{21.1} & 19.9       & 29.9 & \tbf{32.8} & 30.1       & 39.0 & \tbf{46.4} & 38.9       & 37.1 & \tbf{38.7} & 37.2 \\
cat            & animal    & 67.9 & \tif{65.1} & 67.1       & \tif{66.6} & \tif{65.4} & \tif{66.6} & 68.7       & 81.7 & \tif{80.2} & 82.4       & 84.2 & 83.5       & 85.0       & 74.2 & 75.1       & 74.1 \\
dog            & animal    & 47.8 & 47.8       & \tif{45.2} & \tbf{49.7} & \tbf{49}   & \tbf{50}   & 47.0       & 67.5 & 66.6       & 68.1       & 70.2 & \tif{68.7} & 71.0       & 67.3 & 66.9       & 67.0 \\
horse          & animal    & 52.0 & 51.9       & 51.3       & \tbf{53.3} & \tbf{54}   & 51.8       & \tif{50.8} & 64.5 & \tbf{65.5} & 64.5       & 67.2 & \tbf{68.9} & 67.4       & 70.0 & 69.5       & 69.9 \\
sheep          & animal    & 41.3 & \tbf{46.9} & \tbf{44.8} & \tbf{49.4} & \tbf{48.8} & \tbf{47.1} & 42.2       & 58.1 & \tbf{60.9} & 58.2       & 63.8 & \tbf{68.9} & 63.8       & 56.6 & \tbf{59.8} & 56.2 \\
cow            & animal    & 38.3 & 38.7       & \tif{36.7} & \tbf{43.5} & \tbf{41.4} & \tbf{42}   & 38.9       & 57.9 & \tbf{60.8} & 57.2       & 62.2 & \tbf{65.9} & \tif{61}   & 57.1 & \tbf{60.4} & 57.6 \\
elephant       & animal    & 66.0 & 66.2       & \tbf{67.3} & \tbf{70.3} & \tbf{71.3} & \tbf{69.1} & \tbf{67.1} & 79.9 & 79.6       & 79.3       & 80.3 & 81.3       & 80.1       & 79.6 & 80.4       & 79.2 \\
bear           & animal    & 68.2 & 68.9       & \tbf{69.7} & 68.0       & 68.8       & 67.4       & \tif{66.1} & 79.2 & \tbf{83.3} & 79.0       & 80.5 & \tbf{83.6} & 81.0       & 78.8 & 78.6       & 78.7 \\
zebra          & animal    & 75.9 & 76.2       & \tif{74.3} & 75.9       & \tbf{77}   & \tbf{77.2} & 76.0       & 81.8 & 81.4       & 81.6       & 84.0 & 84.2       & 83.7       & 75.2 & 75.9       & 75.9 \\
giraffe        & animal    & 73.9 & 73.5       & \tif{71}   & 74.9       & 73.7       & \tbf{75.3} & 74.9       & 78.8 & 79.7       & 79.7       & 81.6 & \tbf{82.8} & 82.0       & 79.5 & 79.7       & 79.1 \\ \\
backpack       & accessory & 7.0  & 6.2        & 6.8        & 6.8        & 6.5        & 6.5        & 7.3        & 15.7 & \tif{13.9} & 15.8       & 18.4 & \tif{17.1} & 18.5       & 16.6 & 16.1       & 16.5 \\
umbrella       & accessory & 25.9 & 26.5       & 26.9       & \tbf{29.3} & \tbf{28.3} & \tbf{28.4} & \tbf{27}   & 37.0 & \tbf{38.5} & 37.3       & 42.8 & \tbf{46.7} & 43.4       & 43.0 & 43.4       & 42.7 \\
handbag        & accessory & 4.0  & 3.7        & 3.2        & 4.5        & 4.6        & 4.0        & 4.2        & 10.5 & 10.3       & \tbf{11.8} & 14.0 & 14.7       & \tbf{15.8} & 9.8  & \tif{8.6}  & 10.2 \\
tie            & accessory & 24.0 & \tif{21.6} & 24.9       & 24.0       & \tif{21.9} & 23.8       & 23.9       & 32.3 & 32.7       & \tbf{34}   & 38.7 & 38.2       & \tbf{41.5} & 30.7 & 31.5       & \tbf{32} \\
suitcase       & accessory & 20.1 & \tbf{23.5} & \tbf{22.5} & \tbf{26.7} & \tbf{27.3} & \tbf{26}   & \tbf{21.4} & 35.0 & \tbf{39.8} & 35.9       & 37.2 & \tbf{42.3} & 37.6       & 34.7 & \tbf{39.3} & 35.0 \\}
{\megaTableCaption}{all1}

\megaTable{
frisbee        & sports    & 32.5 & \tbf{42.4} & \tbf{34.6} & \tbf{41}   & \tbf{44.9} & \tbf{42}   & 33.2       & 57.8 & \tbf{61.7} & 57.8       & 63.3 & \tbf{67.7} & 62.9       & 53.0 & \tbf{58.3} & 53.2 \\
skis           & sports    & 8.2  & 7.7        & \tbf{9.5}  & 8.8        & 8.9        & 7.6        & 8.9        & 11.6 & \tbf{13.2} & 12.5       & 41.6 & \tbf{45.5} & 42.0       & 11.6 & 11.6       & 12.0 \\
snowboard      & sports    & 18.5 & 18.4       & \tif{13.5} & 19.4       & 17.7       & 18.9       & 18.6       & 27.4 & 27.2       & 28.2       & 38.3 & 39.2       & \tbf{40.2} & 28.0 & 27.6       & 28.6 \\
sports ball    & sports    & 24.0 & \tbf{26.7} & \tbf{26.6} & \tbf{28.7} & \tbf{27}   & \tbf{28.6} & \tbf{26.3} & 37.2 & \tbf{41}   & \tbf{39.2} & 48.1 & \tbf{52.9} & \tbf{50.2} & 40.8 & \tbf{43.4} & \tbf{42.4} \\
kite           & sports    & 20.9 & \tbf{23.8} & 21.4       & \tbf{25.6} & \tbf{25.7} & \tbf{24.6} & \tif{19.5} & 27.1 & \tbf{31.5} & 27.4       & 54.0 & \tbf{62.3} & 53.6       & 40.1 & \tbf{44.7} & 40.2 \\
baseball bat   & sports    & 6.3  & \tbf{8.7}  & \tbf{10.7} & \tbf{10.6} & \tbf{8.9}  & \tbf{9.7}  & \tbf{9.5}  & 21.4 & \tbf{24.6} & \tbf{27.7} & 35.3 & \tbf{39}   & \tbf{44.4} & 24.1 & 25.0       & \tbf{29.5} \\
baseball glove & sports    & 20.0 & \tbf{24.3} & \tbf{29}   & \tbf{30.7} & \tbf{27.8} & \tbf{31.5} & \tbf{27.1} & 37.5 & \tbf{41.9} & \tbf{43.2} & 42.5 & \tbf{49.4} & \tbf{49}   & 35.0 & \tbf{39.3} & \tbf{41.2} \\
skateboard     & sports    & 28.4 & 29.2       & 28.4       & \tbf{33.2} & \tbf{31.3} & \tbf{31.8} & \tbf{29.6} & 46.0 & \tbf{47.5} & \tbf{47.3} & 56.6 & \tbf{59.3} & \tbf{59.8} & 48.4 & \tbf{50.5} & \tbf{51.2} \\
surfboard      & sports    & 23.2 & \tbf{24.7} & 23.5       & \tbf{29.2} & \tbf{25.8} & \tbf{26.4} & \tbf{25.2} & 38.7 & \tif{37.2} & 38.4       & 51.2 & \tbf{54.2} & 51.0       & 35.8 & \tbf{37.9} & 35.6 \\
tennis racket  & sports    & 45.4 & \tbf{46.8} & \tbf{47.7} & \tbf{49.4} & \tbf{48.5} & \tbf{47.2} & \tbf{47.5} & 62.8 & 62.9       & 62.4       & 67.0 & 67.2       & \tbf{68.6} & 57.2 & \tbf{59.2} & \tbf{60.7} \\ \\
bottle       & kitchen    & 14.1 & 14.7       & 13.5       & \tbf{15.7} & 14.9       & 14.9       & 14.3       & 28.4 & 29.2       & 28.6       & 44.4 & \tbf{47.9} & 44.4       & 30.9 & 31.3       & 30.8 \\
wine glass   & kitchen    & 17.4 & \tif{15.4} & 17.0       & \tbf{18.5} & 17.0       & 17.4       & 17.4       & 28.6 & 29.3       & 29.0       & 42.2 & \tbf{44.9} & 41.8       & 23.8 & 23.7       & 24.2 \\
cup          & kitchen    & 19.8 & 20.3       & \tbf{21.7} & \tbf{22.2} & \tbf{21.2} & \tbf{21.5} & 20.0       & 36.4 & 36.9       & 36.5       & 43.3 & \tbf{44.4} & 43.7       & 34.4 & 34.6       & 34.3 \\
fork         & kitchen    & 8.1  & 8.2        & \tbf{9.8}  & \tbf{9.2}  & 9.0        & 8.9        & 8.4        & 15.4 & 14.8       & 15.7       & 18.4 & 18.1       & 18.8       & 17.3 & 16.6       & 17.4 \\
knife        & kitchen    & 3.5  & 4.0        & \tbf{4.7}  & 3.5        & 4.1        & 3.0        & 3.7        & 8.1  & 7.9        & 8.5        & 11.6 & \tbf{13.2} & 11.8       & 10.4 & \tif{9.1}  & 10.3 \\
spoon        & kitchen    & 4.4  & \tif{3.2}  & 3.9        & \tif{2.9}  & \tif{3.3}  & \tif{2.7}  & 4.4        & 7.3  & 7.6        & 7.3        & 10.7 & 11.3       & 11.1       & 6.6  & 6.6        & 6.5 \\
bowl         & kitchen    & 29.3 & 28.7       & 28.5       & \tbf{30.3} & 29.5       & 29.1       & 29.5       & 41.8 & 41.7       & 41.8       & 47.1 & 46.4       & 47.1       & 39.9 & 39.5       & 39.8 \\ \\
banana       & food       & 21.6 & \tif{20}   & 21.8       & 20.8       & \tif{20.4} & \tif{18.4} & 22.5       & 29.7 & 29.7       & 29.5       & 32.8 & \tbf{34.1} & 32.8       & 28.2 & 28.6       & 28.5 \\
apple        & food       & 13.4 & 13.3       & 13.6       & \tbf{14.6} & \tbf{14.7} & \tbf{14.9} & 13.4       & 22.2 & 21.5       & 22.1       & 25.4 & 25.3       & 25.3       & 22.0 & \tif{20.8} & 22.3 \\
sandwich     & food       & 30.5 & 29.8       & \tif{28.9} & 31.4       & 30.8       & \tbf{33.8} & \tbf{32}   & 39.3 & \tif{36}   & \tif{37.9} & 40.4 & \tif{37.2} & \tif{38.7} & 40.0 & \tbf{41.3} & 40.8 \\
orange       & food       & 21.1 & \tif{19.2} & \tif{17.3} & 21.0       & 20.8       & \tif{19.9} & 21.5       & 26.1 & 25.4       & 25.5       & 32.0 & \tbf{33.2} & 32.0       & 30.1 & 30.4       & 30.1 \\
broccoli     & food       & 24.4 & 24.3       & \tbf{25.8} & \tbf{26.7} & \tbf{27.3} & 24.1       & 23.6       & 34.5 & 34.5       & 34.7       & 36.1 & \tbf{37.1} & 36.7       & 37.2 & \tbf{38.3} & 37.2 \\
carrot       & food       & 12.8 & 12.9       & 13.4       & \tbf{14.9} & \tbf{14}   & \tbf{13.9} & 12.8       & 20.1 & 20.3       & 19.4       & 23.3 & \tbf{26}   & 22.8       & 16.8 & 16.9       & 16.6 \\
hot dog      & food       & 23.9 & \tif{22.8} & 24.4       & 24.2       & 23.9       & \tbf{25.6} & 24.0       & 31.4 & \tbf{34.8} & 31.9       & 33.0 & \tbf{36.9} & 33.2       & 28.3 & \tif{26.9} & 27.6 \\
pizza        & food       & 54.5 & 53.9       & \tif{52.6} & \tbf{56.4} & 54.8       & 55.4       & 53.7       & 61.5 & \tbf{63.1} & 61.8       & 63.0 & \tbf{64.9} & 63.0       & 66.0 & 66.4       & 66.3 \\
donut        & food       & 37.2 & \tbf{39.4} & \tbf{41.5} & \tbf{44.8} & \tbf{42.1} & \tbf{43.1} & 37.9       & 55.8 & \tbf{60.1} & 55.9       & 60.7 & \tbf{66.9} & 60.7       & 43.7 & \tbf{47.3} & 43.7 \\
cake         & food       & 28.1 & 27.5       & 28.4       & 28.7       & 28.0       & 27.1       & 27.7       & 41.8 & 41.7       & 42.4       & 43.1 & 43.3       & 43.5       & 35.9 & 35.1       & 36.2 \\
}{\megaTableCaption}{all2}

\megaTable{
chair        & furniture  & 13.3 & 13.7       & 13.7       & \tbf{15.5} & \tbf{14.9} & \tbf{14.4} & 13.4       & 26.3 & \tbf{27.3} & 26.3       & 31.7 & \tbf{34.1} & 31.9       & 26.4 & \tbf{27.5} & 26.3 \\
couch        & furniture  & 37.2 & 37.6       & 37.9       & 36.8       & 37.9       & 37.4       & \tbf{39}   & 52.2 & \tif{50.3} & 52.1       & 53.6 & \tif{51.8} & 53.5       & 50.3 & 49.7       & 50.2 \\
potted plant & furniture  & 20.2 & 19.5       & \tif{18.7} & 21.2       & 20.9       & 21.0       & 19.2       & 28.7 & 27.9       & 28.4       & 34.7 & 35.1       & 34.0       & 27.2 & 26.3       & 26.6 \\
bed          & furniture  & 52.6 & 52.7       & \tbf{54.3} & \tbf{56.2} & 52.9       & \tbf{55.2} & \tif{51.5} & 63.5 & 63.8       & 63.3       & 64.0 & 64.8       & 63.8       & 60.3 & 59.7       & 60.6 \\
dining table & furniture  & 36.5 & 36.9       & \tbf{37.8} & \tbf{37.8} & \tbf{37.9} & 36.9       & 36.2       & 42.2 & 41.7       & 42.4       & 43.1 & 42.7       & 43.3       & 41.7 & 42.3       & 42.1 \\
toilet       & furniture  & 57.7 & 58.3       & 57.6       & \tbf{64.5} & \tbf{59.4} & \tbf{61}   & 57.7       & 69.8 & 69.7       & 70.1       & 73.6 & 74.4       & 74.0       & 67.4 & 67.8       & 68.3 \\ \\
tv           & electronic & 53.9 & \tbf{56.2} & \tbf{56.6} & \tbf{57.9} & \tbf{57.7} & \tbf{57.4} & 53.8       & 65.2 & 65.8       & 65.5       & 66.0 & \tbf{67.4} & 66.4       & 65.4 & \tbf{67.4} & 65.4 \\
laptop       & electronic & 49.9 & \tif{48.5} & \tif{48}   & 50.1       & 49.5       & 49.6       & 50.6       & 65.9 & \tif{63.7} & 66.2       & 67.2 & \tif{65.7} & 67.5       & 67.3 & 68.0       & 67.3 \\
mouse        & electronic & 18.2 & \tbf{28.3} & \tbf{24.2} & \tbf{31.9} & \tbf{27.5} & \tbf{31.1} & 18.9       & 41.4 & \tbf{46.5} & 41.6       & 46.6 & \tbf{52.1} & 46.8       & 47.4 & \tbf{53.6} & 47.7 \\
remote       & electronic & 9.1  & \tif{7.2}  & 9.7        & 8.6        & \tif{8}    & 8.6        & 8.6        & 16.9 & \tbf{18.2} & 16.3       & 21.7 & \tbf{24.5} & 22.3       & 19.7 & \tbf{21.4} & 19.6 \\
keyboard     & electronic & 40.9 & \tbf{42.5} & \tbf{44.5} & \tbf{45.3} & \tbf{43.8} & \tbf{45}   & 40.4       & 54.9 & 55.4       & 55.1       & 57.4 & 57.5       & 57.2       & 57.1 & 57.4       & 57.0 \\
cell phone   & electronic & 16.9 & 16.5       & 16.4       & \tbf{18.1} & \tbf{18.4} & 17.3       & \tbf{18.1} & 29.1 & \tbf{30.5} & 29.1       & 31.4 & \tbf{34.6} & 31.7       & 27.5 & 27.0       & 27.0 \\ \\
microwave    & appliance  & 46.9 & \tbf{51.9} & \tbf{51.9} & \tbf{52.7} & \tbf{54.4} & \tbf{52.3} & 47.3       & 65.7 & 66.4       & 65.4       & 65.9 & 65.1       & 65.7       & 63.1 & \tbf{65.8} & 63.7 \\
oven         & appliance  & 29.6 & 30.4       & 29.5       & \tbf{33}   & \tbf{31.9} & \tbf{31.7} & 30.2       & 44.3 & 43.7       & 44.3       & 45.6 & 46.0       & 45.4       & 47.0 & \tbf{48.3} & 46.6 \\
toaster      & appliance  & 0.0  & \tbf{3.8}  & 0.0        & \tbf{2.1}  & \tbf{2.9}  & 1.0        & 0.0        & 4.3  & \tbf{11.5} & 4.0        & 6.2  & \tbf{15}   & 5.7        & 12.2 & \tbf{17.6} & 12.0 \\
sink         & appliance  & 29.2 & \tbf{32.5} & \tif{27.6} & \tbf{34.9} & \tbf{33.7} & \tbf{33.6} & 29.6       & 42.6 & 42.9       & 42.5       & 47.7 & \tbf{49.6} & 47.6       & 42.6 & 42.6       & 42.8 \\
refrigerator & appliance  & 41.5 & 40.6       & \tif{37.2} & 41.6       & 40.9       & \tbf{42.8} & 41.3       & 53.2 & \tbf{55.8} & 54.0       & 54.4 & \tbf{56.4} & 55.1       & 56.4 & 56.2       & 56.2 \\ \\
book         & indoor     & 7.2  & 7.9        & 7.7        & \tbf{10.1} & 8.1        & \tbf{9.7}  & 7.0        & 11.9 & \tbf{14.1} & 12.0       & 22.5 & \tbf{29.5} & 22.4       & 10.6 & 11.3       & 11.0 \\
clock        & indoor     & 44.1 & \tbf{47.3} & \tbf{46.6} & \tbf{47.4} & \tbf{47.5} & \tbf{47.7} & 43.8       & 56.5 & \tbf{58.2} & 56.5       & 63.1 & \tbf{64.5} & 63.2       & 58.0 & 58.1       & 58.3 \\
vase         & indoor     & 17.2 & \tbf{19.9} & \tbf{21.7} & \tbf{21.6} & \tbf{20.8} & \tbf{21.1} & 17.9       & 32.2 & 33.1       & 33.0       & 37.8 & 38.7       & 38.5       & 40.0 & \tbf{41.8} & 40.2 \\
scissors     & indoor     & 20.0 & 20.4       & \tif{16}   & \tbf{21.3} & 19.8       & \tif{18.6} & \tif{17}   & 27.1 & \tbf{28.9} & 27.2       & 29.3 & \tbf{32.7} & 30.1       & 27.5 & \tif{25.7} & 27.5 \\
teddy bear   & indoor     & 36.5 & \tif{34}   & \tbf{39.1} & 37.0       & 36.4       & \tbf{38.1} & 35.8       & 50.5 & \tbf{52.6} & 50.4       & 52.9 & \tbf{55}   & 53.0       & 50.5 & 49.7       & 51.2 \\
hair drier   & indoor     & 0.0  & 0.0        & 0.0        & 0.0        & 0.0        & 0.0        & 0.0        & 0.0  & 0.1        & 0.0        & 0.0  & 0.1        & 0.0        & 7.9  & \tif{3.5}  & 7.9 \\
toothbrush   & indoor     & 4.7  & \tif{3.7}  & \tif{3.7}  & 4.3        & \tbf{5.9}  & 3.9        & 4.9        & 7.8  & 8.7        & 8.1        & 13.2 & \tbf{15.9} & 13.9       & 10.3 & 10.9       & 10.0 \\
}{\megaTableCaption}{all3}

\paragraph{\fastrcnn Baseline}
Here we describe our \fastrcnn baseline training procedure. We first train
\fastrcnn out of the box on COCO using the default settings. We then start with
this model, and do another round of \fastrcnn training. This round of training
uses the following settings. We freeze all the convolutional layers (till
\texttt{conv5}). We use 1 image per batch and 500 boxes per image in each batch. We 
scale the image to make the shorter side $688$ pixels long. We do 160K iterations
with a learning rate of 0.001 and 160K iterations with a learning rate of
0.0001. Instead of using the inverse of the standard deviation (of bounding box
targets computed from the dataset) to scale the bounding box regression targets, we scale the targets
using a fixed factor of 10. We do not use flipping while training. Finally, our background boxes also include boxes that overlap with ground truth boxes by less than 10\% (\fastrcnn only uses background boxes that overlap with ground truth by at least 10\%).

\paragraph{Differences in training settings for Local Context Model} (line 693)
The local scene context model was trained with the following settings. We do
80K iterations with a learning rate of 0.001 keeping all pre-trained layers
(upto \texttt{fc7}) fixed, 160K iterations with a learning rate of 0.001
keeping all the convolutional layers (till \texttt{conv5}) fixed and the last
80K iterations with a learning rate of 0.0001 with all the convolutional layers
(till \texttt{conv5}) fixed. In addition to this, we also did not train with
foreground sampling, but instead scaled up the loss on foreground windows by a
factor of 10 and the bounding box regression loss by a factor of 5. These
choices were made in preliminary exploration with various models and kept fixed in all subsequent experiments.

\begin{figure}[!h]
\centering
\includegraphics[width=0.49\linewidth]{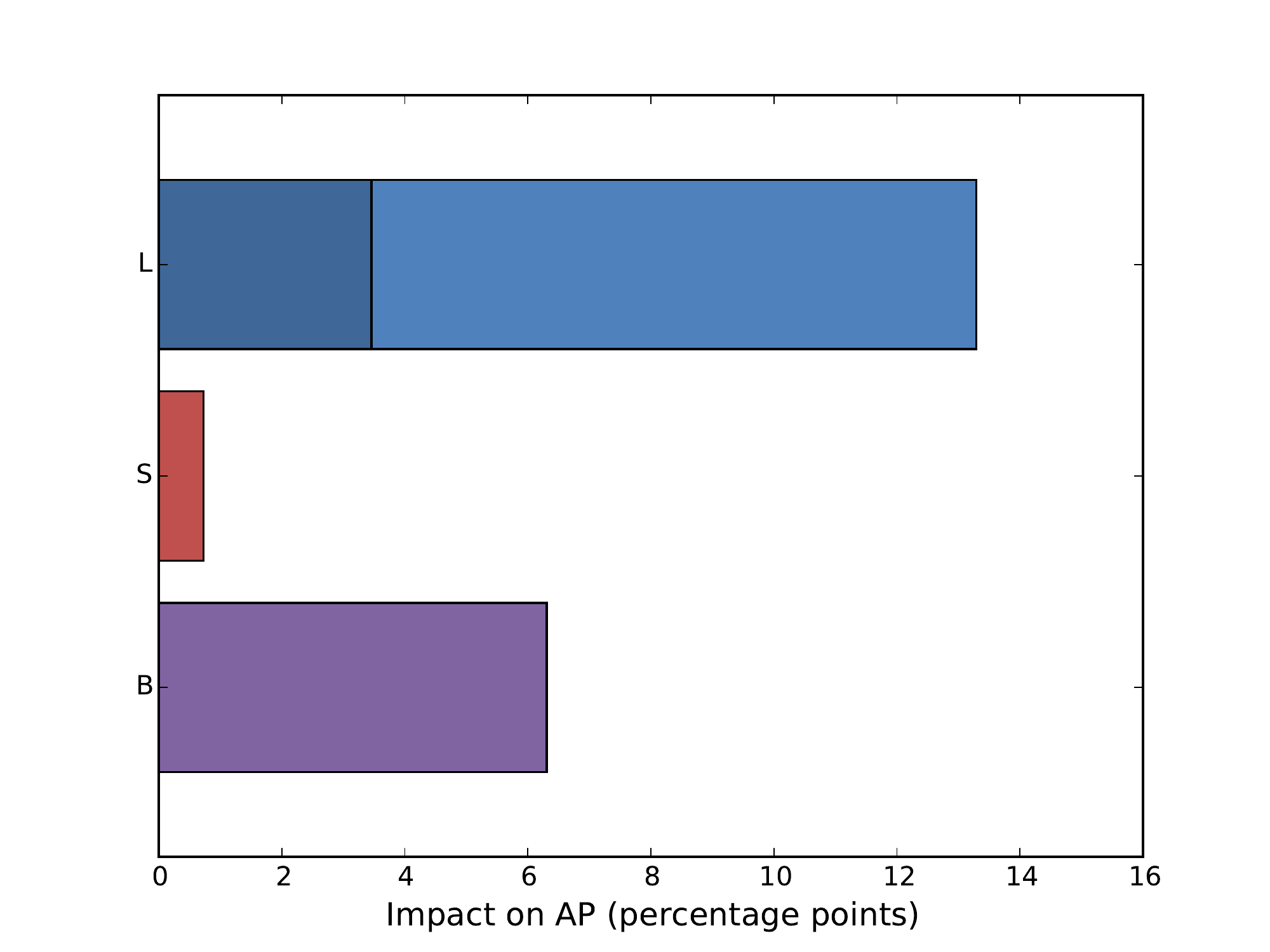}
\includegraphics[width=0.49\linewidth]{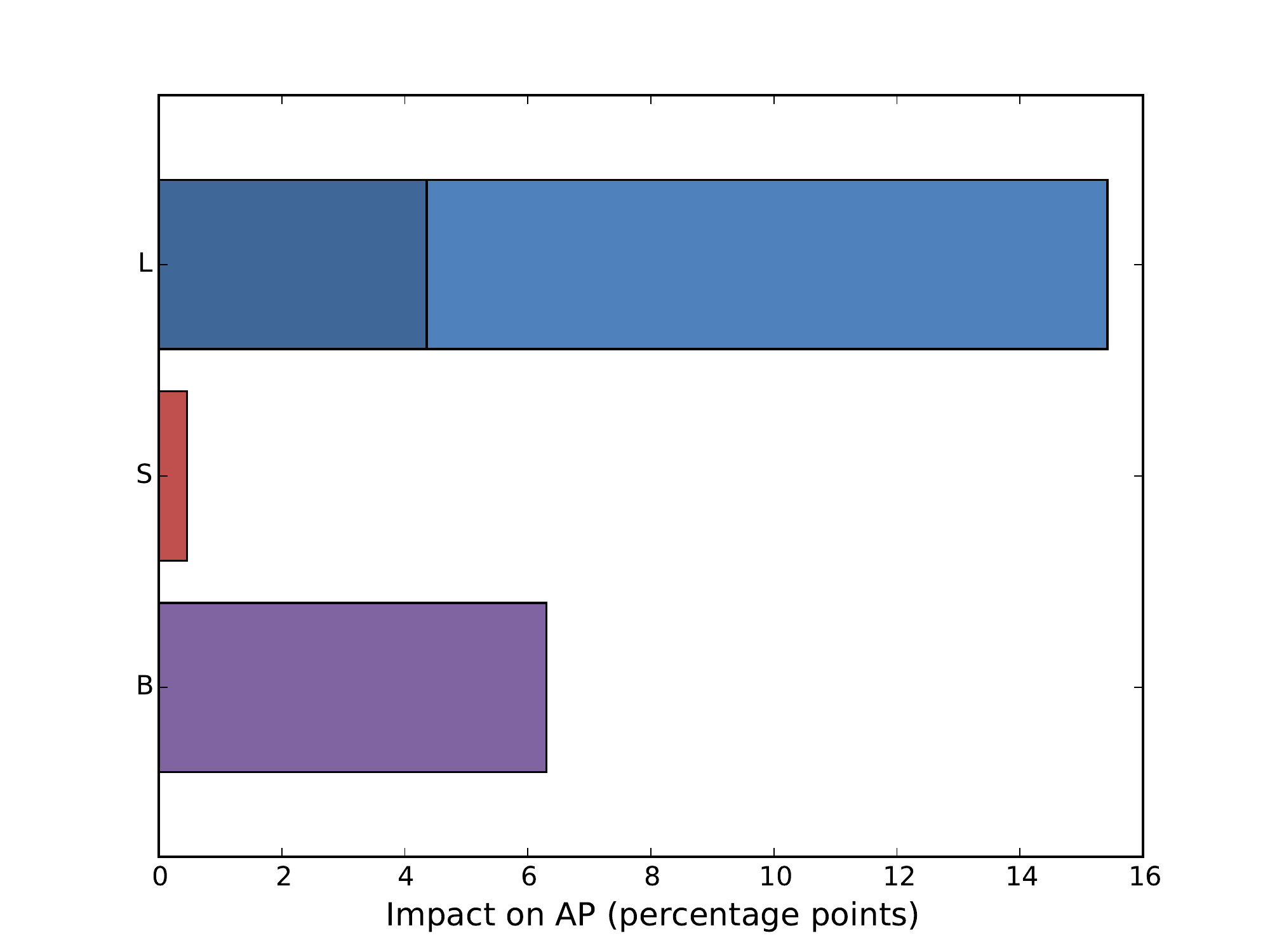}
\caption{Impact on AP due to mislocalization (L), confusion with similar categories (S) and confusion with background (B). Left: \fastrcnn, right: Local Scene Context.}
\label{fig:impact}
\end{figure}

\paragraph{Error analysis}
Figure~\ref{fig:impact} shows a diagnostic plot of how various error modes impact detection performance, for the \fastrcnn baseline and the Local Scene Context model. We measure this impact using the method described by Hoiem et al.~\cite{hoiem2012diagnosing}: we measure how the AP would improve if the respective errors were removed. In particular:
\begin{enumerate}
\item Mislocalized detections include duplicate detections, and detections that overlap a ground truth object by more than 0.1 but less than 0.5. There are two ways of dealing with mislocalization: we can remove mislocalized instances (leading to an improvement in AP shown by the dark blue bar in Figure~\ref{fig:impact}) or we can correct the mislocalization, i.e, improve the overlap of mislocalized instances till it is above 0.5 (leading to an improvement in AP shown by the light blue bar).
\item Sometimes objects of one category are confused with objects of a similar category. To measure the impact of these errors, we use the supercategory groups provided by COCO, and find false positive detections for each category that overlap highly with objects of other categories from the same supercategory. The red bar shows how the AP improves if these false positives are removed.
\item Other false positives are detections firing on background, and the last violet bar indicates how the AP would improve if these false positives were removed.
\end{enumerate}
We find that while the distribution of errors is similar the impact of mislocalization is much higher for the Local Scene Context model. Thus,  compared to \fastrcnn the Local Scene Context model is more prone to fire somewhere in the vicinity of the object, rather than haphazardly on the background. A later localization stage can potentially improve localization and unlock a large gain. In other words, while the Local Scene Context model is slightly better than \fastrcnn at 0.5 overlap, this hides a much larger potential gain if the detections are localized better. 
\paragraph{Per category AP for Table 2 in main paper} We report mean average
precision at 50\% averaged over all classes, averaged over person add-ons, and averaged over each COCO supercategory in \tableref{summary} for all methods reported in
Table 2 in the main paper. We also show the supercategory label for each category. Performance changes of more than 1\% over the
corresponding \fastrcnn baseline are reported in \tbf{blue} (for improvements)
and \tif{red} (for drops). \tableref{all1}, \tableref{all2} and \tableref{all3}
reports individual performance for all categories. 

\balance

\end{document}